\pdfoutput=1
\documentclass[lettersize,journal]{IEEEtran}
\usepackage[T1]{fontenc}       
\usepackage[utf8]{inputenc}    
\usepackage{times}             
\usepackage{latexsym}          
\usepackage{inconsolata}       

\usepackage{verbatim}          
\usepackage{url}               
\usepackage{microtype}         

\usepackage{amsmath, amsfonts, amssymb} 
\usepackage{textcomp}          

\usepackage{array}             
\usepackage{booktabs}          
\usepackage{tabularx}          
\usepackage{multirow}          
\usepackage{makecell}          

\usepackage{graphicx}          
\usepackage{stfloats}          
\usepackage{subcaption}        

\usepackage{algorithmic}       
\usepackage{algorithm}         

\usepackage{cite}              

\begin{document}

\title{Communication-Efficient and Personalized Federated Foundation Model Fine-Tuning via Tri-Matrix Adaptation}
\author{Yongle~Li,  
        Bo~Liu,  
        Sheng~Huang,  
        Zheng~Zhang,  
        Xiao-Tong~Yuan,  
        and~Richang~Hong

\thanks{
This work was supported by the Jointly Funded Project by the National Natural Science Foundation (No. U23B2031). The computation is partially completed on the HPC Platform of Hefei University of Technology.}
\thanks{Yongle Li, Bo Liu, and Richang Hong are with Hefei University of Technology, Hefei, China (e-mail: 2023170713@mail.hfut.edu.cn, kfliubo@gmail.com, hongrc.hfut@gmail.com).}
\thanks{Sheng Huang is with Chongqing University, Chongqing, China (e-mail: huangsheng@cqu.edu.cn).}%
\thanks{Zheng Zhang is with Tsinghua University, Beijing,  China (e-mail: zhengzhang13@icloud.com).}%
\thanks{Xiaotong Yuan is with Nanjing University, Nanjing, China (e-mail: xtyuan1980@gmail.com).}%
}

\markboth{Journal of \LaTeX\ Class Files,~Vol.~14, No.~8, August~2021}%
{Shell \MakeLowercase{\textit{et al.}}: A Sample Article Using IEEEtran.cls for IEEE Journals}


\maketitle

\begin{abstract}
In federated learning, fine-tuning pre-trained foundation models poses significant challenges, particularly regarding high communication cost and suboptimal model performance due to data heterogeneity between the clients. To address these issues, this paper introduces communication-efficient federated LoRA adaption (\textbf{CE-LoRA}), a method that employs a tri-factorization low-rank adaptation approach with personalized model parameter aggregation. We first present a novel LoRA parameter factorization by introducing a small-size dense matrix, which can significantly reduce the communication cost and achieve comparable empirical performance than transferring the low-rank parameter matrix used by existing methods. Without violating data privacy, the server considers the client similarity in both training dataset and model parameter space, and learns personalized weights for model aggregation. Our experiments on various LLM and VLM fine-tuning tasks demonstrate that CE-LoRA not only significantly reduces communication overhead but also improves performance under not independently and identically distributed data conditions. In addition, CE-LoRA improves data privacy protection, effectively mitigating gradient-based data reconstruction attacks. 
\end{abstract}


\section{Introduction}
Pre-trained foundation models (PFMs), such as RoBERTa~\cite{RoBERTa}, GPT~\cite{brown2020language}, and LLaVA~\cite{li2023llavamed,liu2023visual}, are renowned for their stellar performances in a wide range of tasks. With massive training data and well-designed pretraining tasks, these foundational models excel by learning generalizable features, thus enhancing their potential of generalizing to down-stream tasks. Through targeted fine-tuning, these models enhance capabilities in specific domains, such as aligning with human preferences in dialogue systems~\cite{ouyang2022training}, optimizing recommendation algorithms~\cite{lin2024data}, and improving task-specific performance in natural language processing and computer vision~\cite{liu2022ptuning,zhang2021fewshot}.

One pressing challenge for individual clients and small businesses seeking to apply these PFMs to their specific tasks is the lack of domain-specific data for model fine-tuning. Extensive research efforts have been dedicated to optimizing the fine-tuning efficiency of such models~\cite{thangarasa2023spdf,kopf2024openassistant}. In particular, low-rank adaptation (LoRA) and its variants are widely used parameter-efficient fine-tuning methods, which have demonstrated their ability to maintain robust model performance while significantly reducing the number of trainable parameters~\cite{DBLP:conf/iclr/HuSWALWWC22}. More recently, federated learning (FL)-based fine-tuning approaches have been proposed to further address the limitation of insufficient local training data by enabling collaborative model training across multiple clients. The pioneering work FedPETuning~\cite{DBLP:conf/acl/ZhangYDWYQX23} exemplifies how LoRA-based parameter-efficient federated fine-tuning can drastically reduce communication overhead while maintaining adequate performance levels.


\begin{figure}[t]
\centering
\includegraphics[width=1\columnwidth]{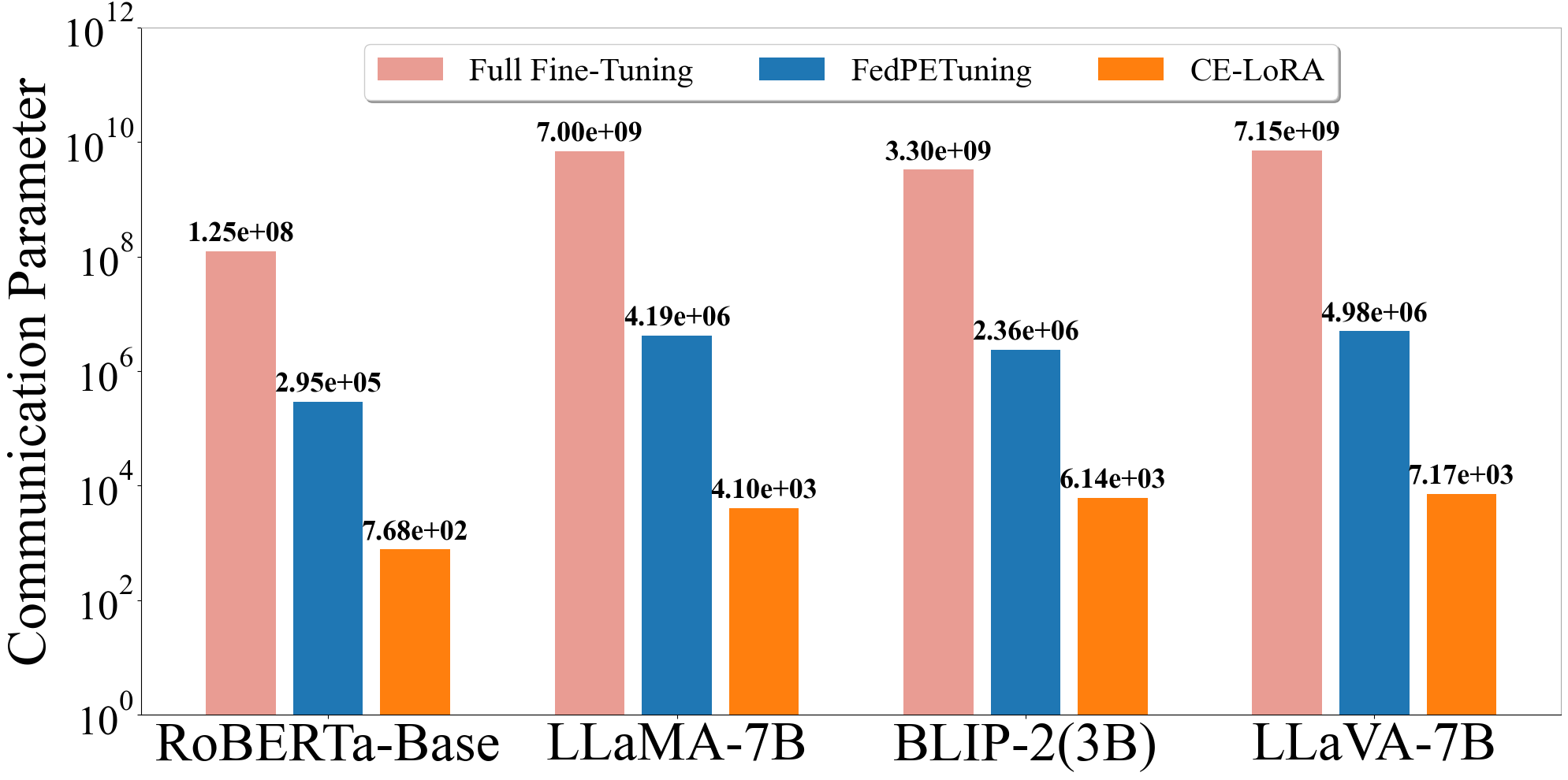} 
\caption{Comparison of communication cost for three federated learning-based fine-tuning methods for pretrained models: transferring all model parameters, LoRA-based FL fine-tuning~\cite{DBLP:conf/acl/ZhangYDWYQX23}, and the proposed CE-LoRA method. The vertical axis is the parameter number needs to be transferred per iteration, on a logarithmic scale. CE-LoRA can reduce communication costs by several hundred times compared to the efficient fine-tuning of LoRA.}
\label{parameters}
\end{figure}

This work aims to address the following two challenges faced by federated fine-tuning of PFMs: (1) As partially illustrated by Figure~\ref{parameters}, even with parameter-efficient fine-tuning techniques like LoRA, federated training of PFMs still requires the transmission of substantial trainable parameters, leading to high inter-client communication costs and low training efficiency; (2) Since data on involved clients tend to be not independently and identically distributed (non-IID), such type of heterogeneity between participants is among the essential technical challenges of FL. The federated average model aggregation strategy~\cite{DBLP:journals/corr/KonecnyMYRSB16,DBLP:conf/aistats/McMahanMRHA17} used in FedPETuning learns a unique global parameter update shared by all participant clients, which fails to consider the training data heterogeneity, leading to sub-optimal performance for each client. 

To address these challenges, our research puts forward a communication-efficient federated LoRA adaptation approach (CE-LoRA), as well as a personalized LoRA parameter aggregation strategy, for PFM fine-tuning. First, within the low-rank-based parameter factorization paradigm of LoRA, we present a triple-parameter factorization by introducing an additional full-rank matrix to vanilla LoRA. The introduced full-rank parameter matrix is updated by inter-client communication in order to encapsulate shared knowledge across all clients. Compared to the two low-rank components in the vanilla LoRA module, the parameter number of this full-rank matrix is much smaller; thus this design can reduce the parameter transmission overloads across clients. 
To achieve effective model personalization, different from the federated average model aggregation used in FedPETuning, we propose to learn a personalized model aggregation for updating each LoRA full rank component on the server. By evaluating the similarity between clients, the proposed aggregation assigns higher weights to the models of clients that are more similar. We propose a client similarity metric that jointly considers dataset similarity and model similarity. With the updated full-rank component by FL, for the two low-rank LoRA matrices, we update them with local fine-tuning. The proposed federated fine-tuning mechanism for PFMs achieves a superior balance between model fine-tuning performance, communication efficiency, and data privacy protection.

We conduct extensive experiments on federated large language model and vision-language model fine-tuning tasks on clients with heterogeneous data to evaluate the performance, communication cost, and privacy protection of CE-LoRA. Compared to state-of-the-art federated LoRA fine-tuning baseline methods, CE-LoRA improves model prediction accuracy while reducing communication overhead by hundreds of times in transmitted parameters. Moreover, CE-LoRA effectively resists gradient-based data reconstruction attacks.

\begin{figure*}[t]
\includegraphics[width=\textwidth]{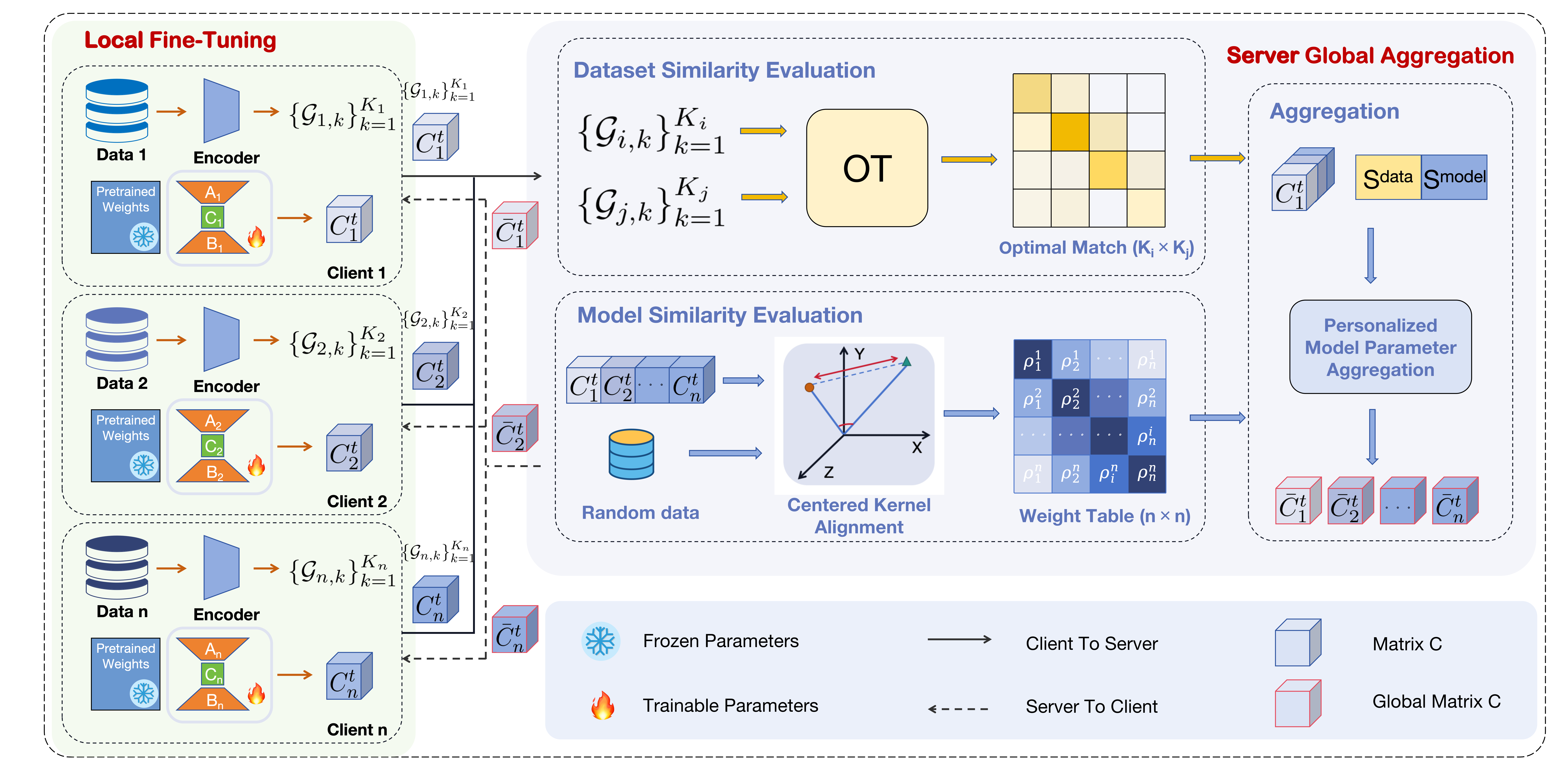}
\caption{Our CE-LoRA framework consists of two main components: local fine-tuning and server global aggregation. In the local fine-tuning stage, we freeze the pretrained foundation model and fine-tune it using  the proposed CE-LoRA method. In addition to the low-rank matrices $A\in \mathbb{R}^{d\times r}, B\in \mathbb{R}^{r\times k}, r\ll \min(k,d)$, we introduce a full-rank matrix $C\in \mathbb{R}^{r\times r}$ that serves as the parameter matrix transmitted between clients and server. After receiving matrices $\{C_{i}^{t}\}_{i=1}^n$, the server calculates the similarity between these parameter matrices using the proposed similarity metric that jointly considers data distribution and model similarity. These client pair-wise similarities are used to derive model aggregation weights for updating $\{\bar{C}_{i}^{t}\}_{i=1}^n$.}
\label{framework}
\end{figure*}

\section{Related Work}
\subsection{Pre-trained Foundation Models}
PFMs have shown broad adaptability across various tasks and domains. These models, typically trained on large-scale data using self-supervised learning, can be fine-tuned to effectively perform a wide range of downstream tasks~\cite{chang2024survey}. According to the model architecture, most existing foundation models can be categorized into encoder-decoder models such as T5~\cite{raffel2020exploring} and BART~\cite{lewis2019bart}, and decoder-only models like the GPT series~\cite{radford2018improving, radford2019language, brown2020language} and LLaMA~\cite{touvron2023llama}. Mixture-of-experts has been used in foundation models as a flexible approach for scaling up model parameters~\cite{fedus2022switch, clark2022unified}.

In recent years, research on PFMs has expanded from single-modal learning to multimodal learning, integrating various data types such as vision and language to improve model understanding~\cite{radford2021learning, liang2024survey}. LLaVA\cite{li2023llava} is a representative vision-language model that combines the LLaMA language model with the CLIP vision encoder and employs alignment tuning for end-to-end training, enabling the model to process image input and perform vision tasks.

Various fine-tuning approaches for foundation models, such as instruction tuning and alignment tuning~\cite{ouyang2022training,tang2024graphgpt, kopf2024openassistant}, have been developed to further adapt these models to specific tasks. Low-rank adapter tuning is one such type of method that avoids full fine-tuning by injecting trainable rank decomposition matrices into each layer of the transformer architecture~\cite{DBLP:conf/iclr/HuSWALWWC22}, inspiring several recent variants. VeRA~\cite{kopiczko2023vera} employs vector-based random matrix adaptation to reduce the number of trainable parameters. AdaLoRA~\cite{zhang2023adalora} adaptively assigns the rank of parameter matrices during fine-tuning. Q-LoRA~\cite{dettmers2024qlora} further reduces the memory overhead of LoRA by incorporating quantization techniques, enabling more efficient fine-tuning without compromising performance. Furthermore, a sparse low-rank adaptation method is proposed in~\cite{ding2023sparse}, while mixture-of-LoRA architectures are explored in~\cite{wu2024mixture, feng2024mixture}.

\subsection{Federated Learning} 
Federated learning is an emerging approach that addresses local training data insufficiency by leveraging data from multiple clients, while preserving data privacy~\cite{longrise,chen2024feddat,bao2024prompt}. Recent advances in federated fine-tuning of large foundation models are presented in works such as~\cite{DBLP:conf/acl/ZhangYDWYQX23,wu2024fedbiot}. In~\cite{guo2024fedlfc,guo2024fedhlt}, clustering-based and tree-based methods are proposed for federated multilingual modeling. A federated multilingual neural machine learning algorithm is introduced in~\cite{liu2023communication}. Several federated prompt tuning algorithms for large models have been proposed in~\cite{che2023federated,dong2023tunable}. Open-source frameworks have recently been developed to support research in federated learning for large models~\cite{ye2024openfedllm,kuang2024federatedscope}. Related topics such as algorithm privacy and efficiency are receiving increasing attention~\cite{vu2024analysis,han2024fedsecurity,qin2023federated,li2024federated}.

Addressing the challenge of data heterogeneity is the primary goal of personalized FL (PFL)~\cite{tan2022towards}. Models such as FedProx~\cite{li2020federated} and MOON~\cite{li2021model} alleviate the impact of data distribution heterogeneity to model with proximal regularization for global model learning. Another category of PFL methods decomposes the model based on parameter update strategies, in which global aggregation or local personalized training is applied to parameter update of different model parts~\cite{ma2022layer,shen2022cd2,wang2024towards}. Differential privacy is applied to federated LoRA learning in~\cite{sun2024improving}. The communication cost of federated LoRA learning is alleviated by sparse pruning in~\cite{kuo2024sparsity}. Recent advances in personalized federated learning for foundation models involve fine-tuning dual LoRA modules~\cite{qi2024fdlora,yang2024dual,wang2024flora}, but a significant challenge arises from the substantial communication costs associated with aggregating LoRA parameters.


\section{Methodology}
\subsection{Overview}                   
An illustration of the proposed CE-LoRA method for federated LoRA-based PFM fine-tuning task is shown in Figure 2. The framework typically contains one server for model aggregation and multiple clients ($\{P_i\}_{i=1}^m$) where each individual client owns its private dataset and model fine-tuning infrastructure. Each client maintains a PFM whose parameters are denoted by $W$. The task is to jointly fine-tune the models by solving the minimization problem
\begin{equation}
    \small
    \min_{\{L_{i}\}_{i=1}^m} \frac{1}{m} \sum_{i=1}^{m} f_i(L_i| W, D_i)
\end{equation}
where $f_i$ denotes the local model training loss for the $i$-th client $P_i$, which is optimized by fine-tuning the LoRA parameters $L_i$ and freezing the PFM parameters; $D_i$ is the local dataset owned by $P_i$.

Algorithm 1 outlines a high-level description of the proposed CE-LoRA algorithm. In the following parts we will elaborate the algorithm details, including the proposed LoRA triple factorization, personalized model parameter aggregation, local fine-tuning and model inference. 

\subsection{LoRA Triple Factorization}
\label{ssect:LTF}
LoRA is among the popular parameter-efficient fine-tuning paradigms, which fine-tunes a product of two low-rank matrices as an auxiliary to the dense parameter matrix. Assume the pre-trained model weight matrix is $W\in \mathbb{R}^{d\times k}$, LoRA learns two low-rank parameter matrices $A\in \mathbb{R}^{d\times r}, B\in \mathbb{R}^{r\times k}$, where $r\ll \min(d, k)$. Given input $x\in \mathbb{R}^d$, the forward pass of that layer output is obtained by
\[
h=x^\top\cdot W+x^\top\cdot A \cdot B.
\]

Although LoRA significantly reduces the number of trainable parameters compared to fine-tuning the entire PFM, the total number of LoRA parameters throughout the entire model can still be substantial, as partially illustrated in Figure~\ref{parameters}. Given the iterative nature of the FL algorithm through parameter communication between the server and the clients, the communication cost of transferring LoRA parameters remains high.

\begin{figure}[t]
\centering
\includegraphics[width=0.75\columnwidth]{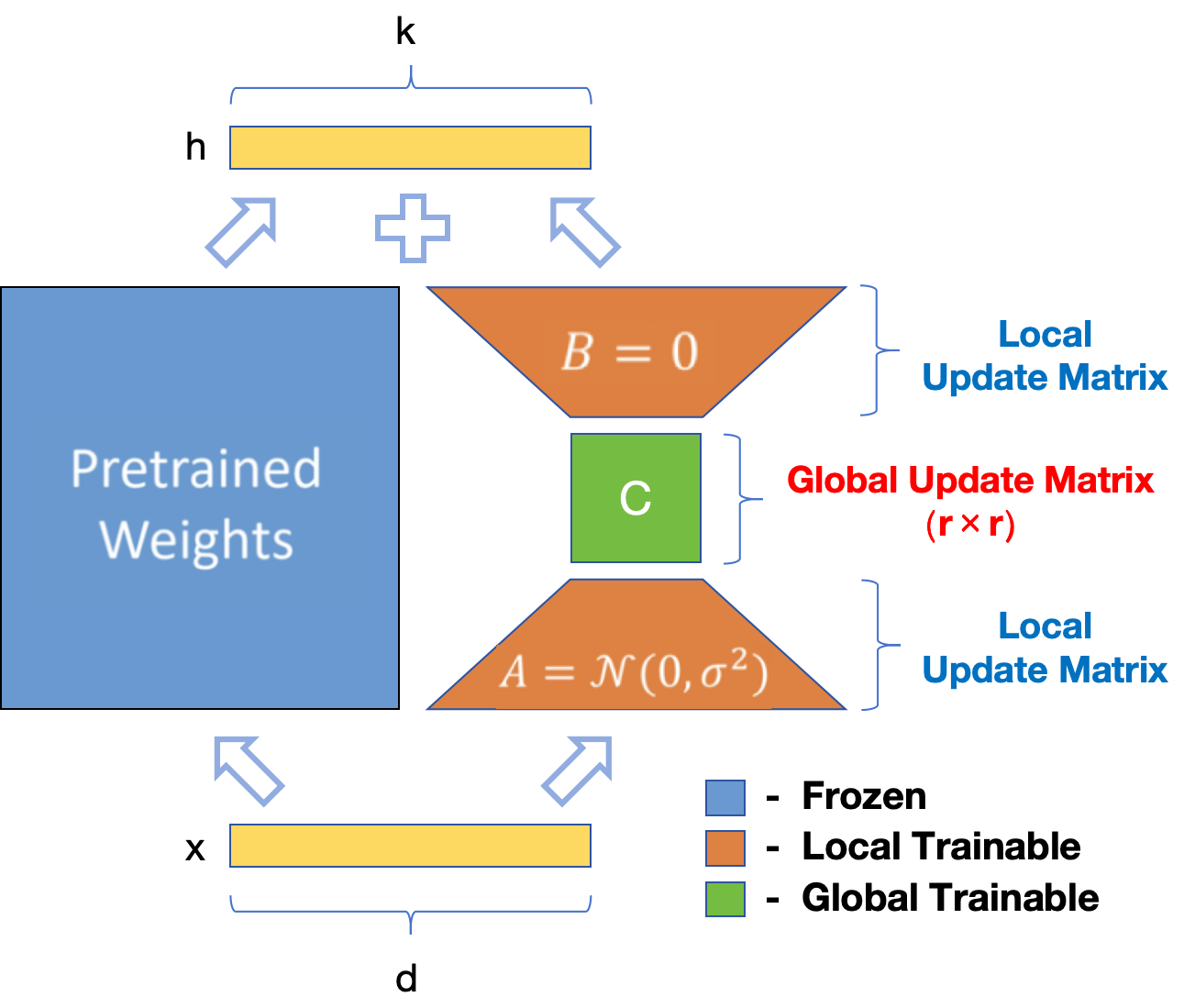} 
\caption{Illustration of the LoRA Triple Factorization. The pre-trained model is frozen during training, while the trainable LoRA is decomposed into \(A \in \mathbb{R}^{r \times d}\), \(B \in \mathbb{R}^{k \times r}\), and \(C \in \mathbb{R}^{r \times r}\), where \(r \ll \min(k, d)\). During federated learning, only \(C\) is transmitted for model parameter aggregation.
}
\label{fig:CE-LoRA}
\end{figure}
In this work, we adopt a triple parameter matrix factorization, which introduces a full-rank matrix $C\in \mathbb{R}^{r\times r}$, in addition to $A$ and $B$ for LoRA. Figure~\ref{fig:CE-LoRA} illustrates the proposed LoRA triple factorization method. With the triple factorization, the LoRA forward pass becomes
\[
h=x^\top\cdot W+x^\top\cdot A \cdot C \cdot B
\]

Instead of transferring both $A$ and $B$ in FedPETuning~\cite{DBLP:conf/acl/ZhangYDWYQX23}, or transferring $B$ only in FFA-LORA~\cite{sun2024improving}, we only transfer $C$ to the server for model parameter aggregation, and update $A$ and $B$ on the local client. Since $r$ is much smaller than $k$ and $d$, using $C$ as the global parameter module, which is updated by aggregation on the server can significantly reduce the per-iteration communication cost between clients and server.

\begin{algorithm}[t]
\caption{Training process of CE-LoRA}
\begin{algorithmic}[1]
\item[\textbf{Inputs:}] Client set $\{P_i\}_{i=1}^m$; Communication round $T$; The pre-trained model parameters $W$; Tri-LoRA parameters $A_i$, $B_i$, $C_i$ for $L_i, i=1,...,m$; The local dataset $D_i$ on client $P_i$.
\FOR{$t = 1$ to $T$}
    \FOR{each \textit{client} $P_i \in \{P_j\}_{j=1}^m$ in parallel}
       \STATE Fine-tune local LoRA to learn $A_{i}^t, C_{i}^t, B_{i}^t$ via
\[ 
\min\limits_{A_i,C_i,B_i}f_i(A_i,C_i,B_i|W,\bar{C}_i^{t-1},D_i);\]
\STATE Send $C_i^t$ to the server;
\STATE Keep $A_i^t$ and $B_i^t$ locally for next round use.
    \ENDFOR
    \STATE \textit{Server} receives  $\{C_{i}^t\}_{i=1}^m$ from the clients;
    \STATE \textit{Server} conducts personalized model parameter aggregation to calculate $\{\bar{C}_{i}^{t}\}_{i=1}^m$ for the clients (refer to \S\ref{ssect:PMPA}); 
    \STATE \textit{Server} send $\{\bar{C}_{i}^{t}\}_{i=1}^m$ to the corresponding clients.
\ENDFOR
\item[\textbf{Outputs:} $\{A_i,C_i, B_i\}_{i=1}^m$.]
\end{algorithmic}
\end{algorithm}

\subsection{Personalized Model Parameter Aggregation}
\label{ssect:PMPA}
Many existing federated PFM fine-tuning methods adopt the federated averaging strategy for global model aggregation. For example,  at the $t$-th iteration, FedPETuning~\cite{DBLP:conf/acl/ZhangYDWYQX23} aggregates the local LoRA parameter matrix $A$ and $B$ on server side with

\[
\bar{A}^t=\frac{\sum\limits_{i=1}^mn_iA_{i}^t}{n}, \quad \bar{B}^t=\frac{\sum\limits_{i=1}^mn_iB_{i}^t}{n},\] 
where $n_i$ and $n$ denote the training sample number of the $i$-th client and the total number of training samples in all clients, respectively. 
However, in FL where data distribution across clients is typically non-IID, the averaging strategy tends to overlook this type of data heterogeneity. 

To alleviate this issue, we design a personalized LoRA parameter aggregation strategy. Instead of learning a unique global model parameter update, we conduct client-specific model aggregation with the proposed personalized aggregation approach, that is 
\begin{equation}
\bar{C}_{i}^t=\phi_i(C_0^t,C_1^t,...,C_m^t), \forall~ \bar{C}_i^t \in \{\bar{C}_j^t\}_{j=1}^m
\label{eqn:agg}
\end{equation}
where $\phi_i$ refers to the personalized aggregation for deriving $\bar{C}_{i}^t$.
It assigns higher weights to models from clients with greater similarity to client $i$.
We formulate Eqn.~\ref{eqn:agg} as
\begin{equation}
\bar{C}_{i}^t = \sum_{j=1,j\ne i}^m \frac{ S_{ij} }{\sum_{j=1,j\ne i}^m S_{ij} } \cdot C_j^t
\end{equation}
where $S_{ij}$ represents the affinity between clients $i$ and $j$.

In this work, we propose a client similarity evaluation metric, which is computed as  
\begin{equation}
S_{ij} = S_{ij}^{\text{data}} + S_{ij}^{\text{model}}
\end{equation}
where $S_{ij}^{\text{data}}$ and $S_{ij}^{\text{model}}$ denotes training dataset similarity and model similarity, respectively. 

\subsubsection{Training Dataset Similarity Evaluation Between Two Clients} 
In FL, directly accessing and comparing raw data between clients poses privacy concerns. To address this, we first learn each client's data distribution using Gaussian Mixture Models (GMM).
GMM is a probabilistic generative model that represents the data distribution as a mixture of Gaussian components~\cite{Reynolds2009}. Suppose that there are $K_i$ categories in the training dataset on client $i$, for category-$k$ of the training dataset, we learn a GMM model with the encoder module output, that is
\[
\mathcal{G}_{i,k}=\sum_{g=1}^Gw_{g}^{(i,k)} \mathcal{N}(\mu_{g}^{(i,k)}, \Sigma_{g}^{(i,k)}) 
\]    
where $G$ is the number of Gaussian component, the model parameters including mixture weights, mean and covariance $\{w_g^{(i,k)}, \mu_g^{(i,k)}, \Sigma_g^{(i,k)}\}_{g=1}^G$ of the Gaussian distributions are learned by EM-type algorithm. We learn the GMMs for all training data categories and obtain $\mathcal{G}_i=\{\mathcal{G}_{i,k}\}_{k=1}^{K_i}$. The client sends all parameters of $\mathcal{G}_i$ to the server to evaluate the similarity of training datasets.

Optimal transport (OT) is a mathematical framework for quantifying the distance between probability distributions by determining the most efficient transport plan that minimizes the total cost of moving the mass from one distribution to another~\cite{peyre2019computational}. We adopt OT to measure the training dataset similarity between $\mathcal{G}_i$ and $\mathcal{G}_j$, which is:  
\begin{equation}\small
S_{ij}^{\text{data}} = \sum_{c,d} \gamma_{cd}^* GW_{cd}
\end{equation}
where \( \gamma_{cd}^* \) denotes the weight of the optimal matching plan between $\mathcal{G}_i$ and $\mathcal{G}_j$, and \(GW_{cd}\) is the Wasserstein-type distance between two GMMs $\mathcal{G}_{i,c}$ and $\mathcal{G}_{j, d}$~\cite{delon2020wasserstein}.

After computing the Wasserstein distances for all category pairs between $\mathcal{G}_{i}$ and $\mathcal{G}_{j}$, we construct the category-level distance matrix \(GW\in \mathbb{R}^{K_i\times K_j}\).
Then we solve the OT problem to find the optimal matching between the two GMM sets via solving  
\begin{equation}\small
\gamma_{cd}^* = \arg\min_{\gamma \in \Pi(\mathcal{G}_i, \mathcal{G}_j)} \sum_{c,d} \gamma_{cd} GW_{cd}
\end{equation}
where $\Pi(\mathcal{G}_i, \mathcal{G}_j)$ denotes the set of all joint distributions with marginals $\mathcal{G}_i$, $\mathcal{G}_j$. It can be efficiently solved using the Sinkhorn algorithm with entropy regularization.

It is notable that the training dataset similarity evaluation is a one-shot effort. We compute the pairwise similarities of the clients at the beginning of the FL process, then use the result in the FL model learning iterations.

\subsubsection{Training Model Similarity Evaluation Between Two Clients} 
In addition to the training dataset diversity across the clients, the model fine-tuning mechanism differences  respect to hyper-parameter setting, optimizer etc. can also bring heterogeneity to the model from the involved clients. To quantize this type of heterogeneity, we evaluate the model similarity $S_{ij}^{\text{model}}$, which is calculated using the Centered Kernel Alignment (CKA) method~\cite{kornblith2019similarity}. We define  
\begin{equation}
S_{ij}^{\text{model}} = \text{CKA}(C_i^t, C_j^t).
\end{equation}

CKA was originally used to assess the similarity of representations between models or model layers. The computation of CKA relies on the Hilbert-Schmidt Independence Criterion (HSIC), which is used to measure the dependency between two sets of variables. Specifically, to evaluate the similarity between $C_i^t$ and $C_j^{t}$, we first randomly generate a set with $n$ samples, which then pass through by $C_i^{t}$ and $C_j^{t}$, respectively. We use $K_i^{t}\in \mathbb{R}^{n\times n}$ and $K_j^{t}\in \mathbb{R}^{n\times n}$ to denote the linear kernel of the output.  The CKA-based similarity metric is calculated via
\begin{equation}
\text{CKA}(C_i^t, C_j^t) = \frac{\text{HSIC}(C_i^t, C_j^t)}{\sqrt{\text{HSIC}(C_i^t, C_i^t) \cdot \text{HSIC}(C_j^t, C_j^t)}}
\end{equation}
where $\text{HSIC}(C_i^t, C_j^t)$ is computed by:
\begin{equation}
\text{HSIC}(C_i^t, C_j^t) = \operatorname{tr}(K_i^{t}HK_j^{t}H)
\end{equation}
where $\operatorname{tr}$ denotes the matrix trace, and $H=I-\frac{1}{n}\mathbf{1} \mathbf{1}^\top$ is the centering matrix.

\subsection{Local Fine-tuning and Model Inference}
For each of the clients $P_i$, upon receiving a global parameter update $\bar{C}_i^{t-1}$ from the server, it utilizes its local data $D_i$ to fine-tune its LoRA parameters $A_i$, $C_i$ and $B_i$ by solving
\[ 
\min\limits_{A_i,C_i,B_i}f_i(A_i,C_i,B_i|W,\bar{C}_i^{t-1},D_i).\]
The client freezes the parameters of the pre-trained model $W$, and initializes $A$, $C$ and $B$ with $A_i^{t-1}$, $\bar{C}_i^{t-1}$ and $B_i^{t-1}$, respectively. After reaching the convergence of model fine-tuning, the client sends $C_i^t$ to the server for a new iteration round. 

Once the federated PFM fine-tuning completes, for each layer that requires computation, the original pre-trained model weight matrix \( W \) is adjusted by the product of the trained low-rank matrices \( A_i \), \(C_i\), and \( B_i \), that is 
\begin{equation}
    W_i = W + A_i \cdot C_i \cdot B_i
\end{equation}
where \( W_i \) is the fine-tuned weight matrix used for inference for client $P_i$.

\section{Experiments}

\subsection{Experiments Setup}

\noindent\textbf{Models and Datasets:} 
We evaluate four representative models spanning natural language processing and computer vision domains. For NLP tasks, we employ RoBERTa-base (125M parameters), an enhanced BERT variant optimized for language understanding, and LLaMA-7B (7B parameters), a large language model specialized for text generation and reasoning. For vision-related tasks, we utilize BLIP-2 (3B parameters), a vision-language model with two-stage image-text alignment training, and LLaVA-7B which integrates visual encoders with LLaMA-7B for multimodal understanding. All models employ LoRA adaptation with default rank 8.

Our experiments use six benchmarks spanning text and image domains. For NLP datasets, SST-2 and MNLI from the GLUE benchmark are adopted for sentiment analysis and textual entailment respectively, following~\cite{kairouz2021advances} in repurposing their original validation sets as test sets; AG\_NEWS serves as a news classification benchmark. For computer vision datasets, we employ CIFAR-10, CIFAR-100 and ImageNet as benchmarks. For CIFAR-10 and CIFAR-100, which lack validation sets, we create validation sets by splitting their training data. For ImageNet, we used the official validation set for evaluation. Table~\ref{tab:dataset} summarizes the training, validation, and testing sample sizes for these datasets.


\begin{table}[t] 
\centering
\resizebox{0.7\linewidth}{!}{
\begin{tabular}{ccccc}
\toprule
Dataset & Train & Dev & Test  \\
\midrule
SST-2 & 66,675 & 674 & 872  \\
MNLI & 388,774 & 3,928 & 9,815  \\
AG\_NEWS & 120,000 & 7,600 & 7,600  \\
CIFAR-10 & 40,000 & 10,000 & 10,000  \\
CIFAR-100 & 40,000 & 10,000 & 10,000  \\
ImageNet-1K & 1,231,167 & 50,000 & 50,000  \\
\bottomrule
\end{tabular}
}

\caption{Benchmark datasets used in the experiments.}
\label{tab:dataset}
\vspace{-0.4cm}
\end{table}

\noindent\textbf{Non-IID Data Partition on Clients:}
We follow the previous works~\cite{DBLP:conf/acl/ZhangYDWYQX23,wang2023fedabc} that utilize the Dirichlet distribution \(\text{Dir}(\alpha)\) to partition the dataset, in which a smaller \(\alpha\) indicates higher data heterogeneity. In our experiments, the data is partitioned across 10 clients, and \(\alpha = 0.5\) as the default setting.


\noindent\textbf{Baselines:}
To validate the effectiveness of CE-LoRA, we compare it with six baselines: (1)  LoRA training with local data; (2) FedPETuning~\cite{DBLP:conf/acl/ZhangYDWYQX23}, which learns vanilla LoRA in a federated approach; (3) FFA-LoRA~\cite{sun2024improving}, an optimized scheme within the federated learning framework that only transmits matrix $B$; (4) FDLoRA~\cite{qi2024fdlora}, a personalized federated method combining client-specific and global LoRA modules through adaptive fusion; (5) pFedMe-LoRA, implementing pFedMe's Moreau envelope~\cite{t2020personalized} with full LoRA parameter aggregation; (6) pFedMe-FFA, a variant of pFedMe that adopts FFA-LoRA's communication mechanism to transmit only matrix $B$.

\begin{table*}[t]
\centering
\small

\resizebox{\textwidth}{!}{
\begin{tabular}{c|c|ccc||c|c|ccc}
\toprule
\multicolumn{5}{c||}{\textbf{NLP Benchmarks}} & \multicolumn{5}{c}{\textbf{CV Benchmarks}} \\
\cmidrule(lr){1-5} \cmidrule(lr){6-10}
Model & Method & \makebox[1.2cm]{SST-2} & \makebox[1.2cm]{MNLI} & \makebox[1.2cm]{AG\_NEWS} & Model & Method & \makebox[1.2cm]{CIFAR-10} & \makebox[1.2cm]{CIFAR-100} & \makebox[1.2cm]{ImageNet} \\
\midrule
\multirow{7}{*}{RoBERTa} & LoRA-Loc     & 91.2 & 80.9 & 79.1 & \multirow{7}{*}{BLIP-2} & LoRA-Loc     & 81.1 & 56.2 & 66.2 \\
                         & FedPETuning  & 91.1 & 83.4 & 80.2 &                         & FedPETuning  & 82.4 & 59.6 & 66.3 \\
                         & FFA-LoRA     & 91.7 & 84.2 & 81.9 &                         & FFA-LoRA     & 82.9 & 59.8 & 68.2 \\
                        & pFedMe-LoRA    & 92.3 & 85.4 & 82.6 &                         & pFedMe-LoRA   & 83.5 & 60.9 & 69.3 \\
                        & pFedMe-FFA    & 92.4 & 85.2 & 82.5 &                         & pFedMe-FFA    & 83.3 & 60.5 & 69.1 \\
                        & FDLoRA     & 92.8 & 85.6 & 82.6 &                         & FDLoRA     & 83.7 & 61.1 & 69.4 \\
                         & CE-LoRA      & \textbf{93.2} & \textbf{86.1} & \textbf{83.5} & & CE-LoRA      & \textbf{84.4} & \textbf{62.5} & \textbf{70.1} \\
\midrule
\multirow{7}{*}{LLaMA}   & LoRA-Loc     & 94.3 & 89.1 & 82.3 & \multirow{7}{*}{LLaVA}  & LoRA-Loc     & 85.2 & 64.2 & 70.4 \\
                         & FedPETuning  & 96.2 & 91.2 & 88.7 &                         & FedPETuning  & 85.8 & 66.1 & 71.3 \\
                         & FFA-LoRA     & 96.2 & 91.7 & 90.0 &                         & FFA-LoRA     & 86.1 & 66.5 & 73.2 \\
                        & pFedMe-LoRA    & 96.3 & 92.4 & 90.8 &                         & pFedMe-LoRA   & 86.8 & 67.6 & 75.4 \\
                        & pFedMe-FFA    & 96.2 & 92.2 & 90.5 &                         & pFedMe-FFA    & 86.4 & 67.5 & 75.2 \\
                        & FDLoRA     & 96.1 & 92.8 & 91.0 &                         & FDLoRA     & 87.0 & 67.6 & 75.9 \\
                         & CE-LoRA      & \textbf{96.4} & \textbf{93.1} & \textbf{91.5} & & CE-LoRA      & \textbf{87.2} & \textbf{68.8} & \textbf{76.6} \\
\bottomrule
\end{tabular}
}
\caption{Accuracy comparison of fine-tuning methods on NLP and CV benchmarks. Left: RoBERTa and LLaMA on NLP datasets. Right: BLIP-2 and LLaVA on CV datasets.}
\label{merged_exper}
\vspace{-0.4cm}
\end{table*}

\begin{table}[t]
\centering
\small
\resizebox{0.95\linewidth}{!}{
\begin{tabular}{c|c|c|c}
\toprule
Domain & Model & Method & \makecell{Com.\\ (Percentage)} \\
\midrule
\multirow{12}{*}{NLP} 
& \multirow{6}{*}{RoBERTa} 
& FedPETuning & \(2.95 \times 10^5\) (100\%) \\
&& pFedMe-LoRA & \(2.95 \times 10^5\) (100\%) \\
&& FDLoRA & \(2.95 \times 10^5\) (100\%) \\
&& pFedMe-FFA & \(1.47 \times 10^5\) (50\%) \\
&& FFA-LoRA & \(1.47 \times 10^5\) (50\%) \\
&& CE-LoRA & \(\mathbf{7.68 \times 10^2}\) (0.26\%) \\
\cmidrule{2-4}
& \multirow{6}{*}{LLaMA} 
& FedPETuning & \(4.19 \times 10^6\) (100\%) \\
&& pFedMe-LoRA & \(4.19 \times 10^6\) (100\%) \\
&& FDLoRA & \(4.19 \times 10^6\) (100\%) \\
&& pFedMe-FFA & \(2.10 \times 10^6\) (50\%) \\
&& FFA-LoRA & \(2.10 \times 10^6\) (50\%) \\
&& CE-LoRA & \(\mathbf{4.10 \times 10^3}\) (0.10\%) \\
\specialrule{.1em}{.5em}{.5em}
\multirow{12}{*}{CV} 
& \multirow{6}{*}{BLIP-2} 
& FedPETuning & \(2.36 \times 10^6\) (100\%) \\
&& pFedMe-LoRA & \(2.36 \times 10^6\) (100\%) \\
&& FDLoRA & \(2.36 \times 10^6\) (100\%) \\
&& pFedMe-FFA & \(1.18 \times 10^6\) (50\%) \\
&& FFA-LoRA & \(1.18 \times 10^6\) (50\%) \\
&& CE-LoRA & \(\mathbf{6.14 \times 10^3}\) (0.26\%) \\
\cmidrule{2-4}
& \multirow{6}{*}{LLaVA} 
& FedPETuning & \(4.98 \times 10^6\) (100\%) \\
&& pFedMe-LoRA & \(4.98 \times 10^6\) (100\%) \\
&& FDLoRA & \(4.98 \times 10^6\) (100\%) \\
&& pFedMe-FFA & \(2.49 \times 10^6\) (50\%) \\
&& FFA-LoRA & \(2.49 \times 10^6\) (50\%) \\
&& CE-LoRA & \(\mathbf{7.17 \times 10^3}\) (0.14\%) \\
\bottomrule
\end{tabular}
}

\caption{Communication cost comparison (Com. indicates parameters transmitted per iteration, percentage shows compression ratio relative to FedPETuning)}

\label{merged_comparison}

\vspace{-0.4cm}
\end{table}

\subsection{Performance Comparison}

\begin{figure}[htbp]
    \centering
    \begin{minipage}[t]{0.48\linewidth}
        \centering
        \includegraphics[width=\linewidth]{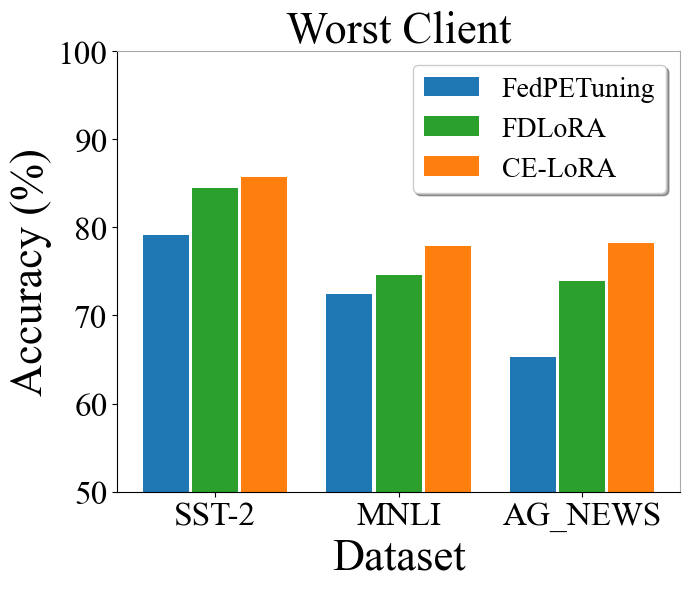}
    \end{minipage}
    \hfill
    \begin{minipage}[t]{0.48\linewidth}
        \centering
        \includegraphics[width=\linewidth]{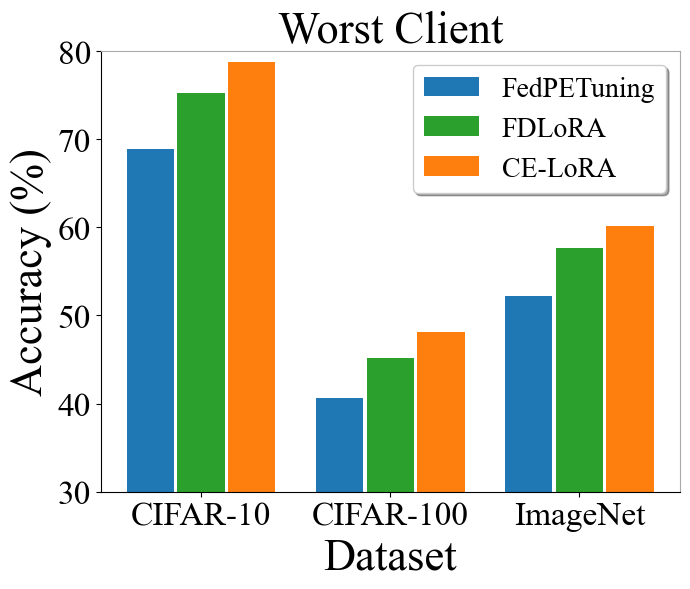}
    \end{minipage}

    \begin{minipage}[t]{0.48\linewidth}
        \centering
        \includegraphics[width=\linewidth]{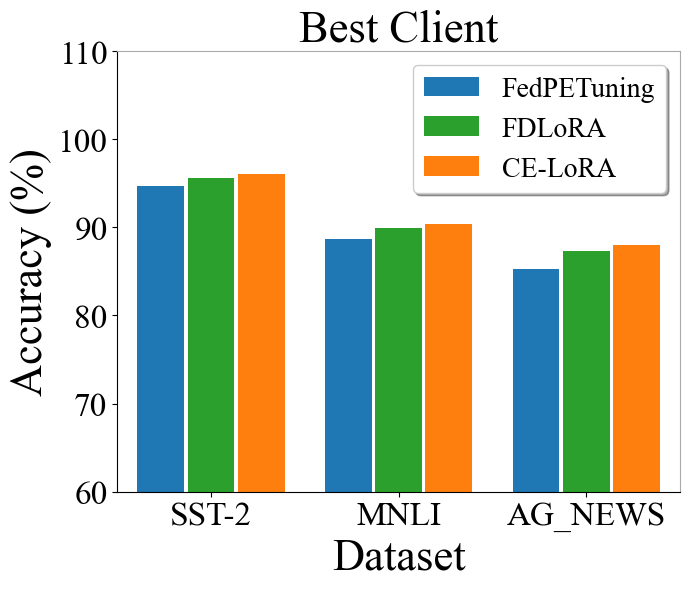}
    \end{minipage}
    \hfill
    \begin{minipage}[t]{0.48\linewidth}
        \centering
        \includegraphics[width=\linewidth]{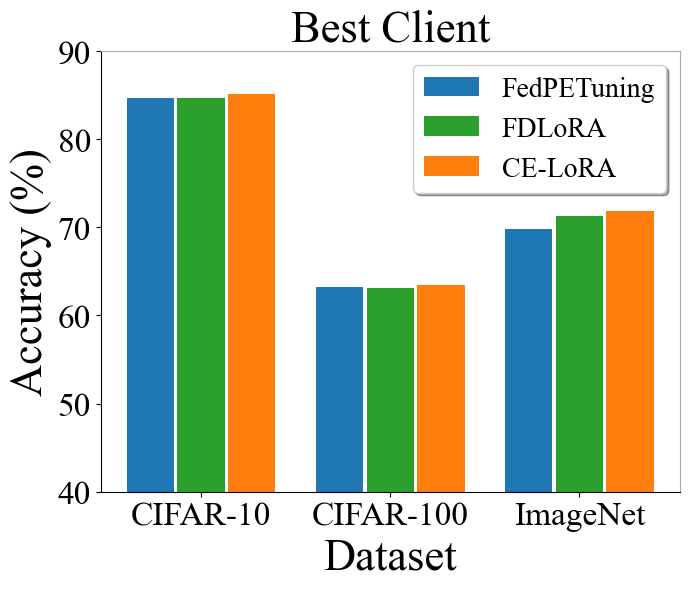}
    \end{minipage}
    
    \caption{Performance comparison of worst-performing client and best-performing client.}
    \label{client}

\vspace{-0.4cm}
\end{figure}

\begin{figure}[h]
    \centering
    \begin{subfigure}[b]{0.15\textwidth}
        \includegraphics[width=\textwidth]{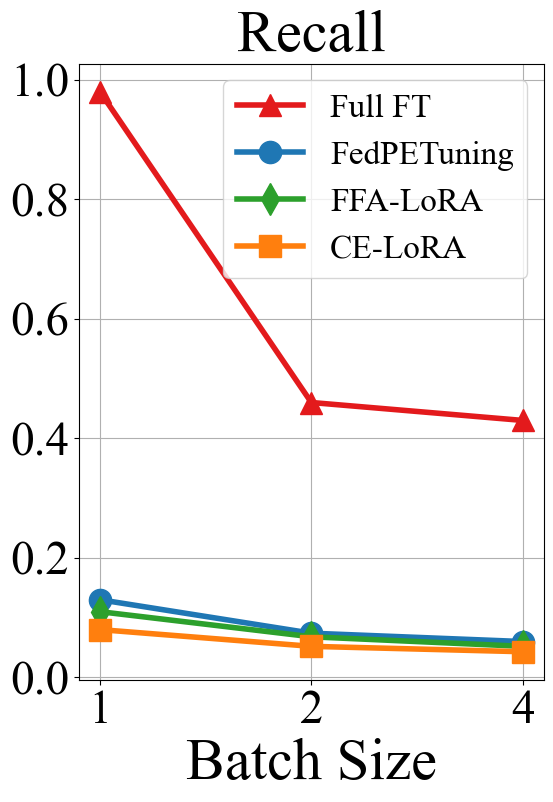} 
        \label{fig:sub1}
    \end{subfigure}
    \hfill 
    \begin{subfigure}[b]{0.15\textwidth}
        \includegraphics[width=\textwidth]{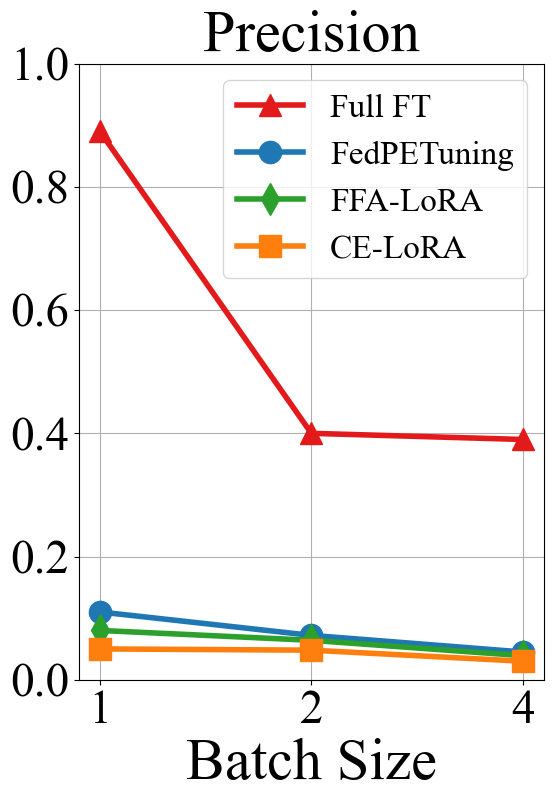}
        \label{fig:sub2}
    \end{subfigure}
    \hfill 
    \begin{subfigure}[b]{0.15\textwidth}
        \includegraphics[width=\textwidth]{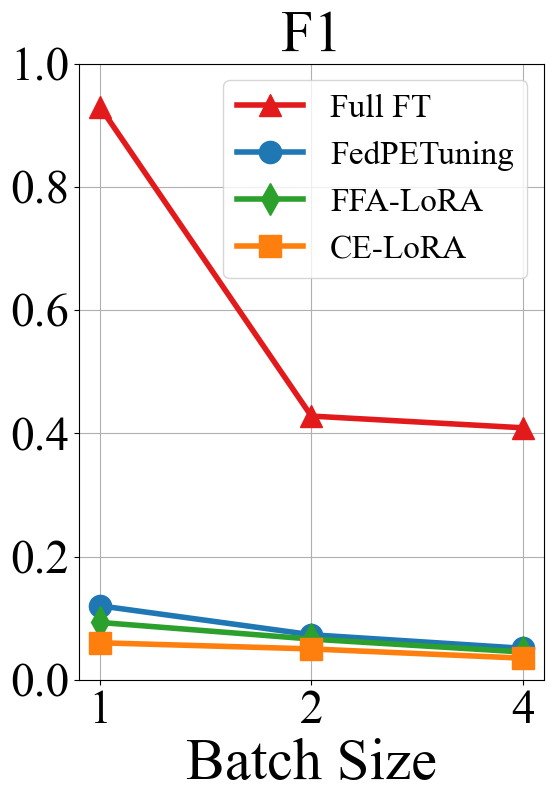}
        \label{fig:sub3}
    \end{subfigure}
    \vspace{-0.3cm}
    \caption{The comparison of the data reconstruction attack on the dataset using full PFM fine-tuning, FedPETuning, FFA-LoRA and CE-LoRA.}
    \label{Privacy}
\vspace{-0.4cm}
\end{figure}


Table~\ref{merged_exper} and Table~\ref{merged_comparison} respectively present the performance and communication cost comparisons of CE-LoRA against the baselines. The results indicate that CE-LoRA enhances the model's ability to adapt to local personalized data and also significantly reducing communication overhead. For example, on the CIFAR-100 dataset, CE-LoRA achieves a relative image classification accuracy improvement of over $2.29\%$ and $1.78\%$ for BLIP-2 and LLaVA model fine-tuning over the second-best algorithm FDLoRA. For communication cost comparison, we use FedPETuning, which transmits all LoRA parameters in each round as baseline, and evaluate the communication cost reduction. For the LLaMA model, CE-LoRA reduces communication overhead by a factor of 1024 compared to FedPETuning and by a factor of 512 compared to FFA-LoRA.

Figure~\ref{client} illustrates the performance of the best and worst clients among 10 clients in the three federated PFM fine-tuning methods. CE-LoRA consistently improves model performance for the considered models and datasets and significantly outperforms the other methods in the worst-performing client. The results indicate that CE-LoRA can effectively improve the model performance for clients with insufficient training data, which depend on FL to help improve the model performance.

\subsection{Privacy Protection Experiment}
For privacy protection evaluation, following~\cite{DBLP:conf/acl/ZhangYDWYQX23}, we employ the DLG method~\cite{zhu2019deep}, an attack technique that reconstructs data from gradients. We measure precision (the average percentage of correctly recovered words in the target text), recall (the average percentage of words in the target text that were correctly predicted) and the F1 score (the harmonic mean of precision and recall). We randomly select 128 samples from MNLI as the attack dataset.

Figure~\ref{Privacy} shows the DLG attack results of four representative methods with different communication transmission matrices. The results indicate that CE-LoRA can defend against data reconstruction attacks more effectively than other fine-tuning methods, and its privacy protection performance is relatively unaffected by batch sizes. This is because CE-LoRA only transmits $r\times r$ parameters of the LoRA model per-iteration, and it is more difficult for attackers to reconstruct data from such a limited number of model parameters.

\subsection{Ablation Study}

In this section, we conducted three ablation experiments on the RoBERTa and LLaVA models to validate the effectiveness of our proposed matrix decomposition and personalized model aggregation method. The experimental results are shown in Table~\ref{ablation_nlp} and Table~\ref{ablation_cv}.
We first compare our triple LoRA factorization method with vanilla LoRA while using federated average as the model aggregation strategy. The result demonstrates that the proposed LoRA factorization can maintain performance with significantly reduced communication overhead. Then we test applying personalized model aggregation over using federated average. Compared to traditional federated average, the proposed personalized global aggregation effectively improves the performance of the model in heterogeneous environments. 

\begin{table}[t]
\centering
\small
\resizebox{\linewidth}{!}{
\begin{tabular}{c|ccc}
\toprule
 Method & \makebox[1.3cm]{SST-2} & \makebox[1.3cm]{MNLI} & \makebox[1.3cm]{AG\_NEWS} \\
 \midrule
LoRA + FedAvg & 91.1 & 83.4 & 80.2  \\
Tri-LoRA + FedAvg & 91.2 & 83.3 & 81.5  \\
Tri-LoRA + $\text{S}^{\text{data}}$ & 92.6 & 83.7 & 81.9  \\
Tri-LoRA + $\text{S}^{\text{data}}$ + $\text{S}^{\text{model}}$ & \textbf{93.2} & \textbf{86.1} & \textbf{83.5} \\
\bottomrule
\end{tabular}
}
\caption{Ablation study experiment results on NLP datasets using RoBERTa. ``Tri-LoRA'' represents the proposed triple LoRA factorization discussed in \S\ref{ssect:LTF}. ``$\text{S}^{\text{data}}$'' and ``$\text{S}^{\text{model}}$'' represent dataset similarity and model similarity for personalized model parameter aggregation, detailed in \S\ref{ssect:PMPA}.}

\label{ablation_nlp}
\vspace{-0.3cm}
\end{table}

\begin{table}[t]
\centering
\small
\resizebox{\linewidth}{!}{
\begin{tabular}{c|ccc}
\toprule
 Method & \makebox[1.3cm]{CIFAR-10} & \makebox[1.3cm]{CIFAR-100} & \makebox[1.3cm]{ImageNet} \\
\midrule
LoRA + FedAvg & 85.8 & 66.1 & 71.3 \\
Tri-LoRA + FedAvg & 85.9 & 66.4 & 71.6  \\
Tri-LoRA + $\text{S}^{\text{data}}$ & 87.1 & 66.9 & 75.8  \\
Tri-LoRA + $\text{S}^{\text{data}}$ + $\text{S}^{\text{model}}$ & \textbf{87.2} & \textbf{68.8} & \textbf{76.6} \\
\bottomrule
\end{tabular}
}

\caption{Ablation study experiment results on CV datasets using LLaVA. Method notations are consistent with Table~\ref{ablation_nlp}.}
\label{ablation_cv}
\vspace{-0.3cm}
\end{table}

\subsection{Performance Comparison under Varying Data Skew}

\begin{figure*}[t]
    \centering
    \begin{subfigure}[b]{0.24\textwidth}
        \includegraphics[width=\textwidth]{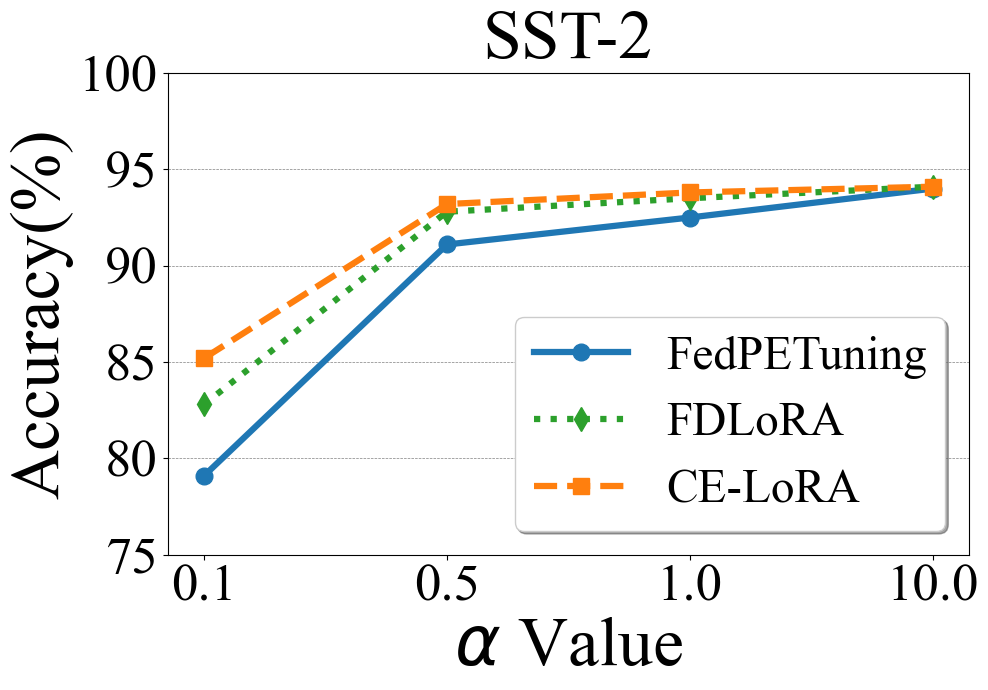} 
    \end{subfigure}
    \hfill 
    \begin{subfigure}[b]{0.24\textwidth}
        \includegraphics[width=\textwidth]{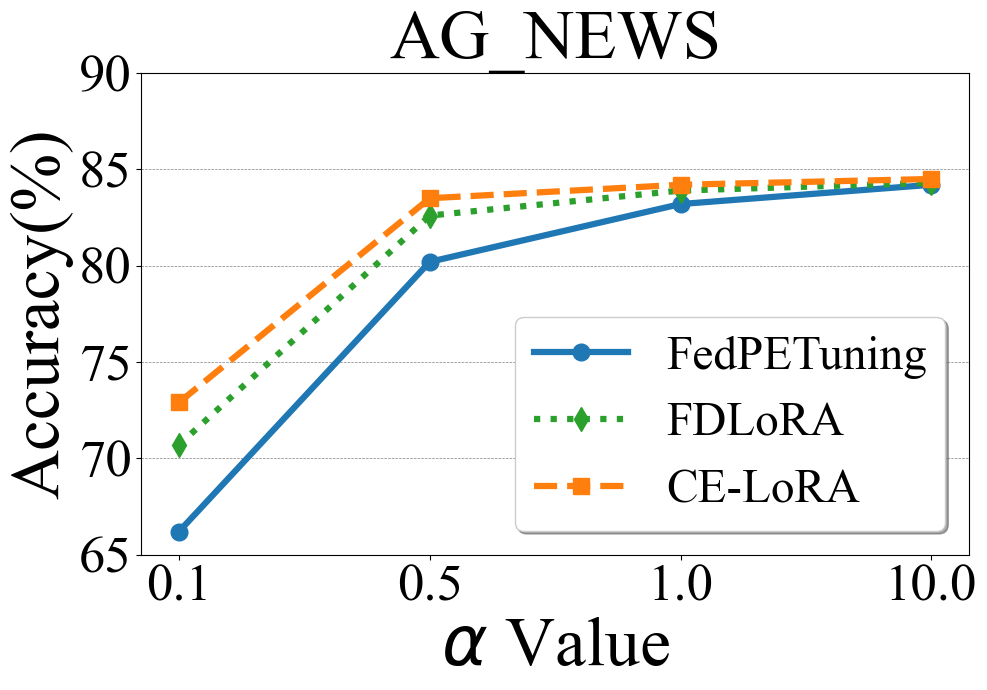}
    \end{subfigure}
    \hfill 
    \begin{subfigure}[b]{0.24\textwidth}
        \includegraphics[width=\textwidth]{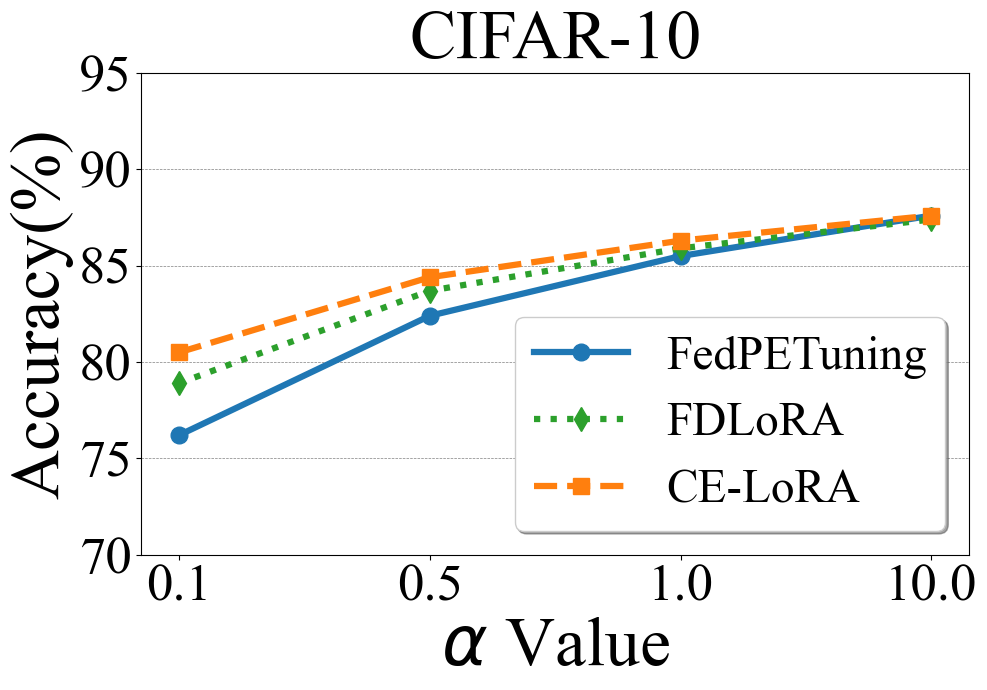}
    \end{subfigure}
    \hfill 
    \begin{subfigure}[b]{0.24\textwidth}
        \includegraphics[width=\textwidth]{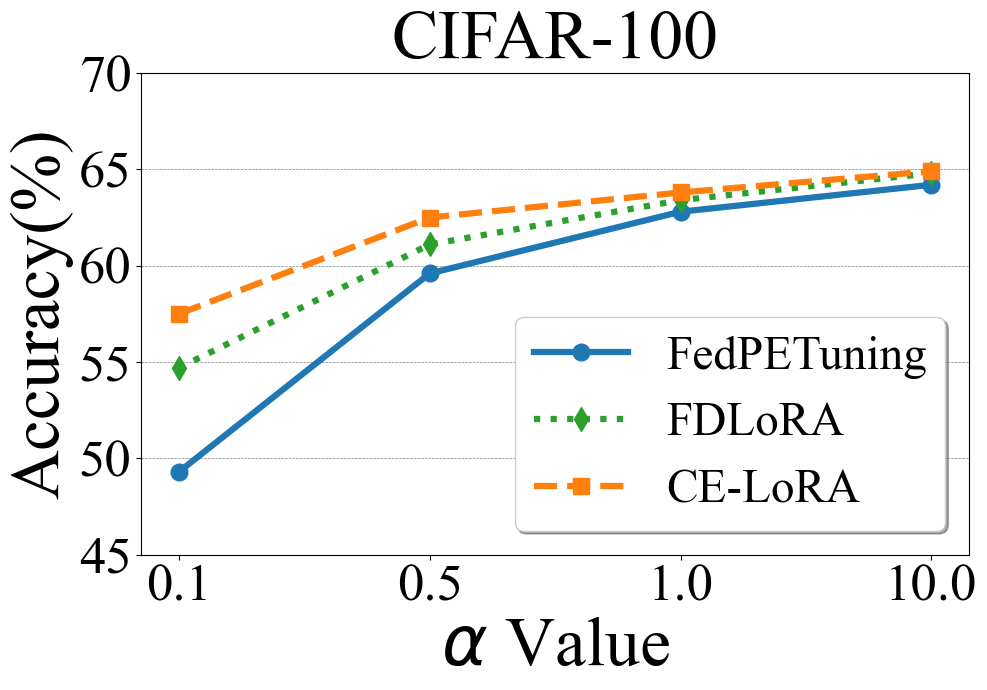}
    \end{subfigure}

\caption{Performance comparison of FedPETuning, FDLoRA and CE-LoRA under different data heterogeneity degree parameterized by varying $\alpha$ values.}
\label{alpha}
\vspace{-0.3cm}
\end{figure*}

\begin{figure}[htbp]
    \centering
    \begin{minipage}[b]{0.48\linewidth}
        \centering
        \includegraphics[width=\linewidth]{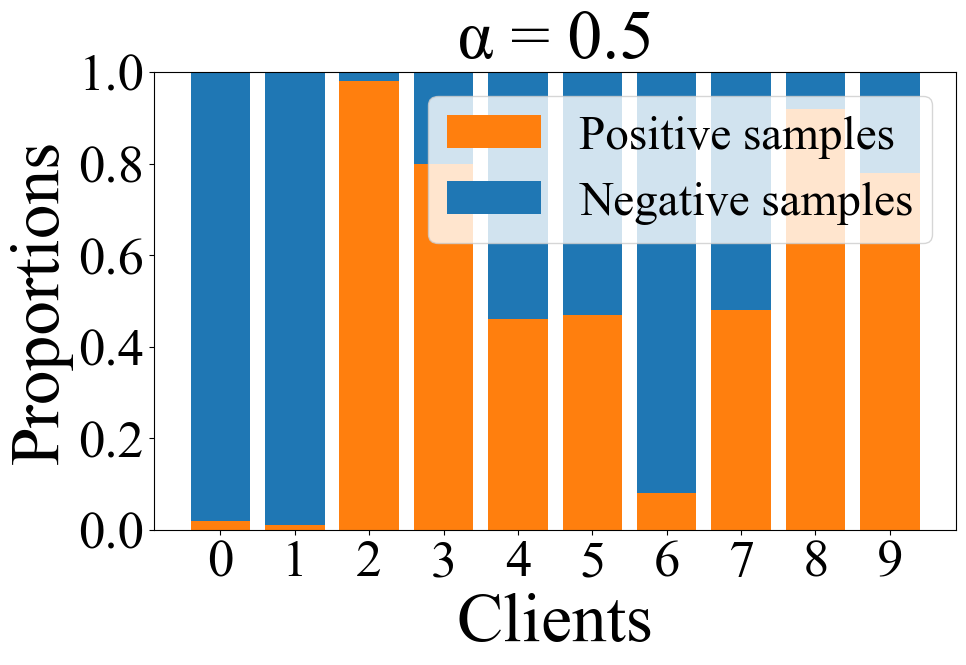}
    \end{minipage}
    \begin{minipage}[b]{0.48\linewidth}
        \centering
        \includegraphics[width=\linewidth]{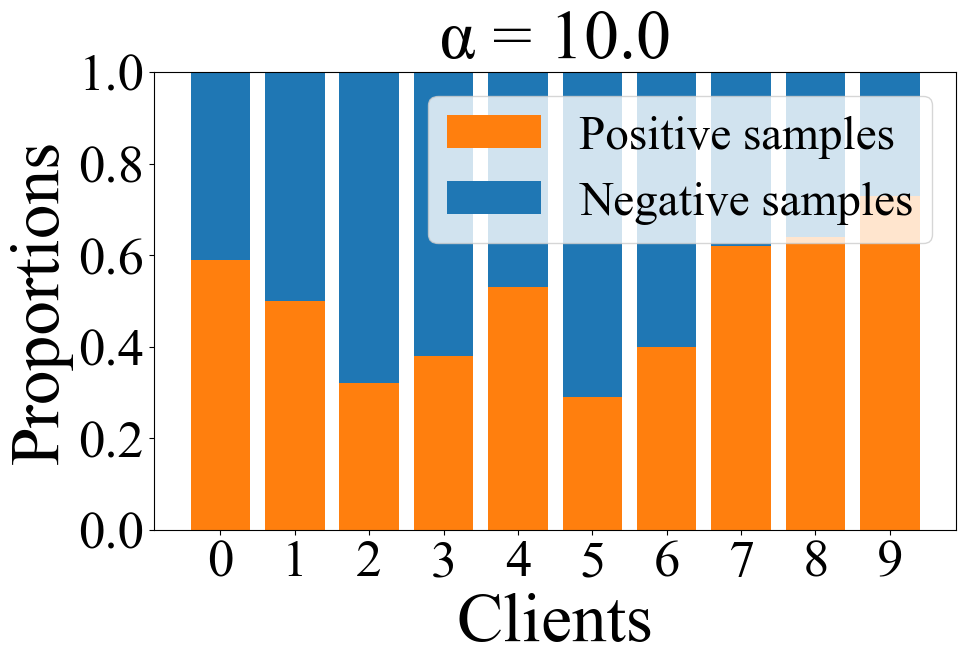}
    \end{minipage}
    \caption{The training set label distribution of the SST-2 dataset under different Dirichlet parameter $\alpha$ for the 10 clients. }
    \label{Heterogeneity}
\vspace{-0.3cm}
\end{figure}

We evaluate FedPETuning, FDLoRA, and CE-LoRA under varying data heterogeneity levels, fine-tuning RoBERTa on SST-2 and AG\_NEWS, and BLIP-2 on CIFAR-10 and CIFAR-100. In this experiment, four Dirichlet distributions are tested by setting $\alpha$ as \{0.1, 0.5, 1.0, 10.0\}. As shown in Figure~\ref{Heterogeneity}, the smaller the $\alpha$, the more heterogeneous data distribution between clients. When $\alpha$ is 10.0, the data across most clients is nearly uniformly distributed.

As shown in Figure~\ref{alpha}, CE-LoRA, FedPETuning, and FDLoRA are found to have a decrease in performance as the value of $\alpha$ gets decreases. However, CE-LoRA is clearly less affected by data heterogeneity. This is because CE-LoRA employs a personalized model aggregation scheme on LoRA component $C$, together with LoRA local fine-tuning for the component $A$ and $B$, which leads to better model generalization to local data. 


\subsection{Performance Comparison under Varying Client Numbers}

\begin{figure}[htbp]
    \centering
    \begin{minipage}[b]{0.48\linewidth}
        \centering
        \includegraphics[width=\linewidth]{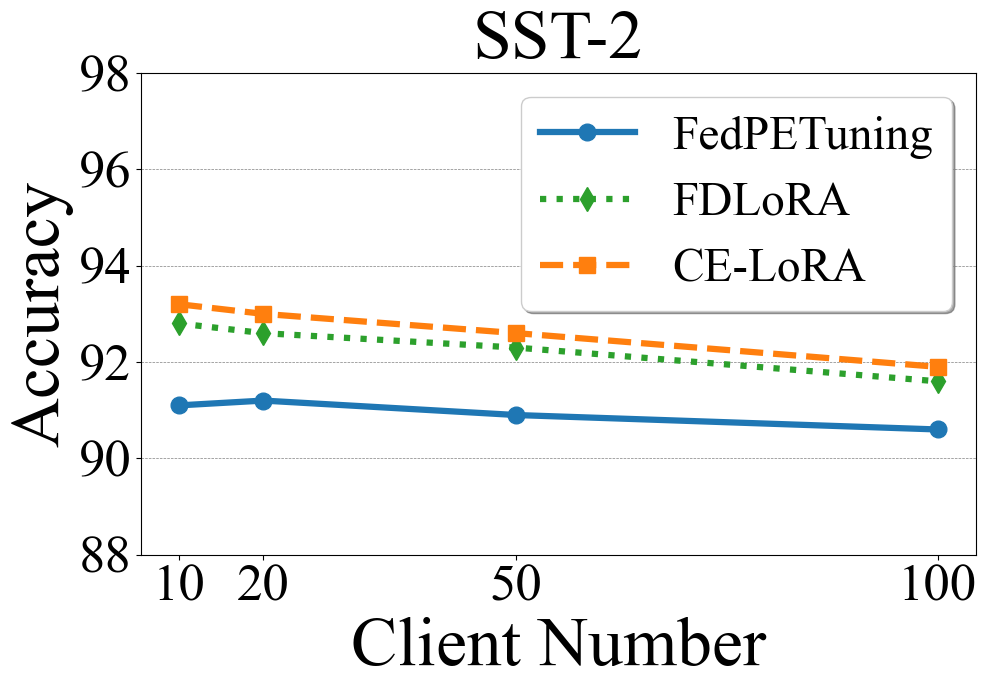}
    \end{minipage}
    \begin{minipage}[b]{0.48\linewidth}
        \centering
        \includegraphics[width=\linewidth]{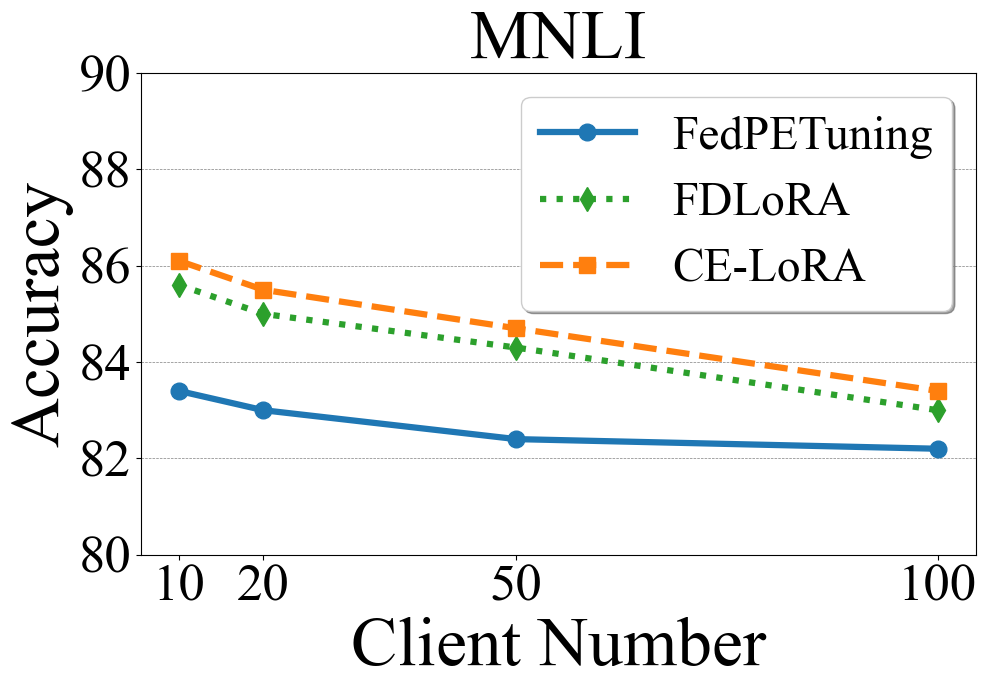}
    \end{minipage}

    \begin{minipage}[b]{0.48\linewidth}
        \centering
        \includegraphics[width=\linewidth]{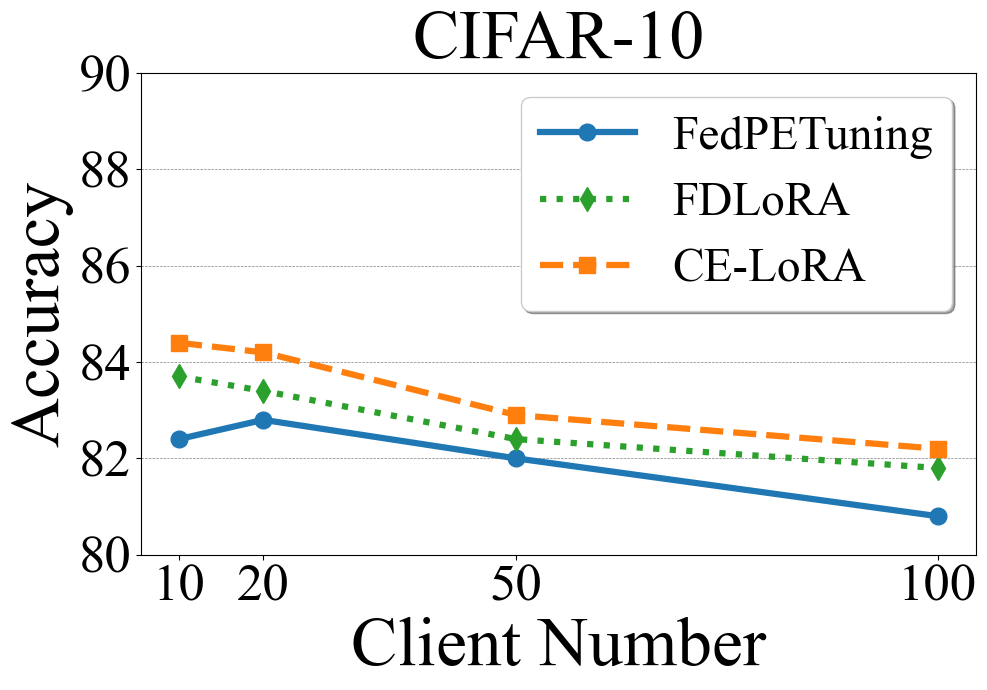}
    \end{minipage}
    \begin{minipage}[b]{0.48\linewidth}
        \centering
        \includegraphics[width=\linewidth]{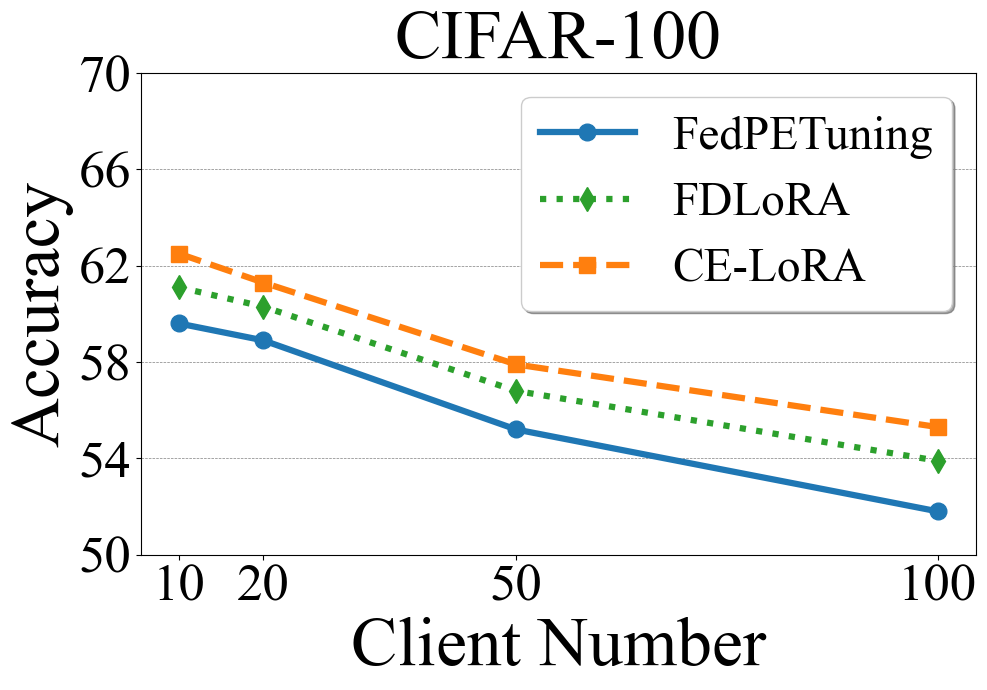}
    \end{minipage}
    
    \caption{Performance comparison of FedPETuning, FDLoRA, and CE-LoRA under varying clients numbers.}
    \label{client_number}
\end{figure}

We compare the performance of FedPETuning, FDLoRA, and CE-LoRA by partitioning the benchmark datasets into a varying number of clients. We fine-tune RoBERTa on SST-2 and MNLI and BLIP-2 on Cifar10 and Cifar100, respectively. The number of clients is set to 10, 20, 50, and 100. As shown in Figure~\ref{client_number}, CE-LoRA consistently outperforms both FedPETuning and FDLoRA as the number of clients increases. 
\begin{figure*}[t]
    \centering
    \begin{subfigure}[b]{0.3\textwidth}
        \includegraphics[width=\textwidth]{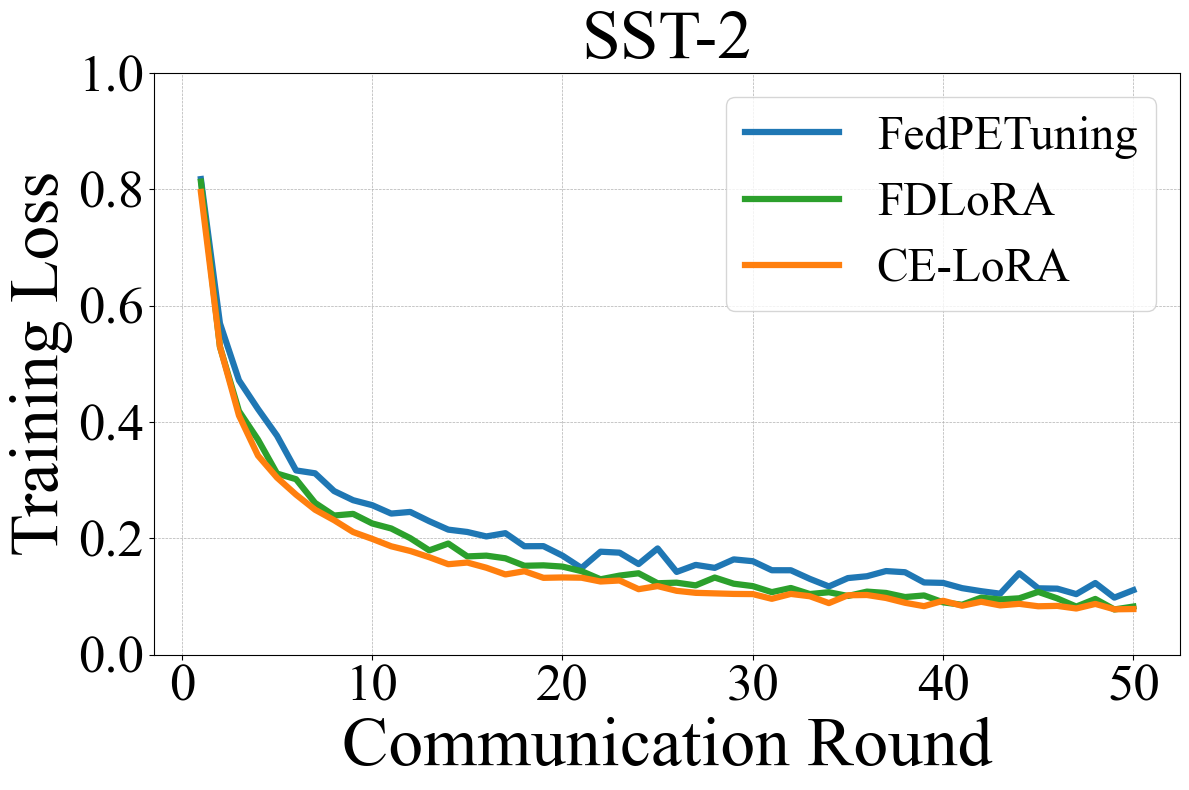} 
    \end{subfigure}
    \hfill 
    \begin{subfigure}[b]{0.3\textwidth}
        \includegraphics[width=\textwidth]{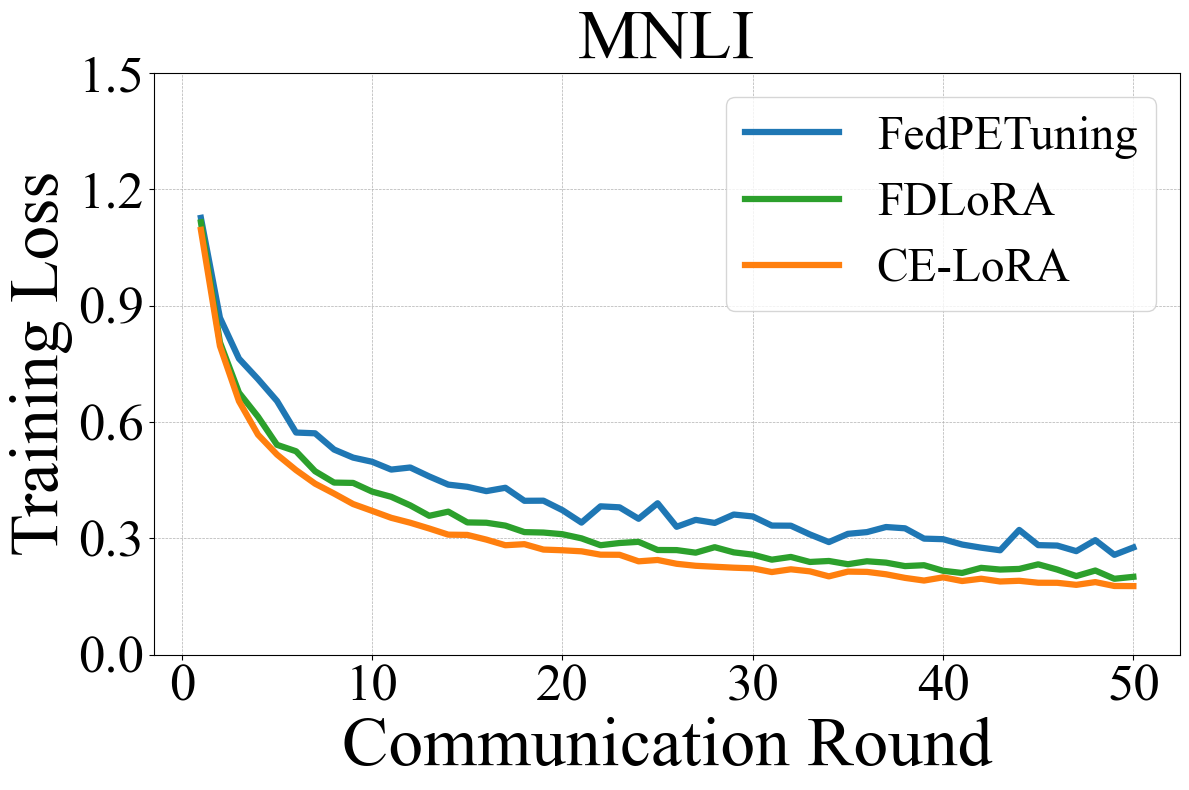}
    \end{subfigure}
    \hfill 
    \begin{subfigure}[b]{0.3\textwidth}
        \includegraphics[width=\textwidth]{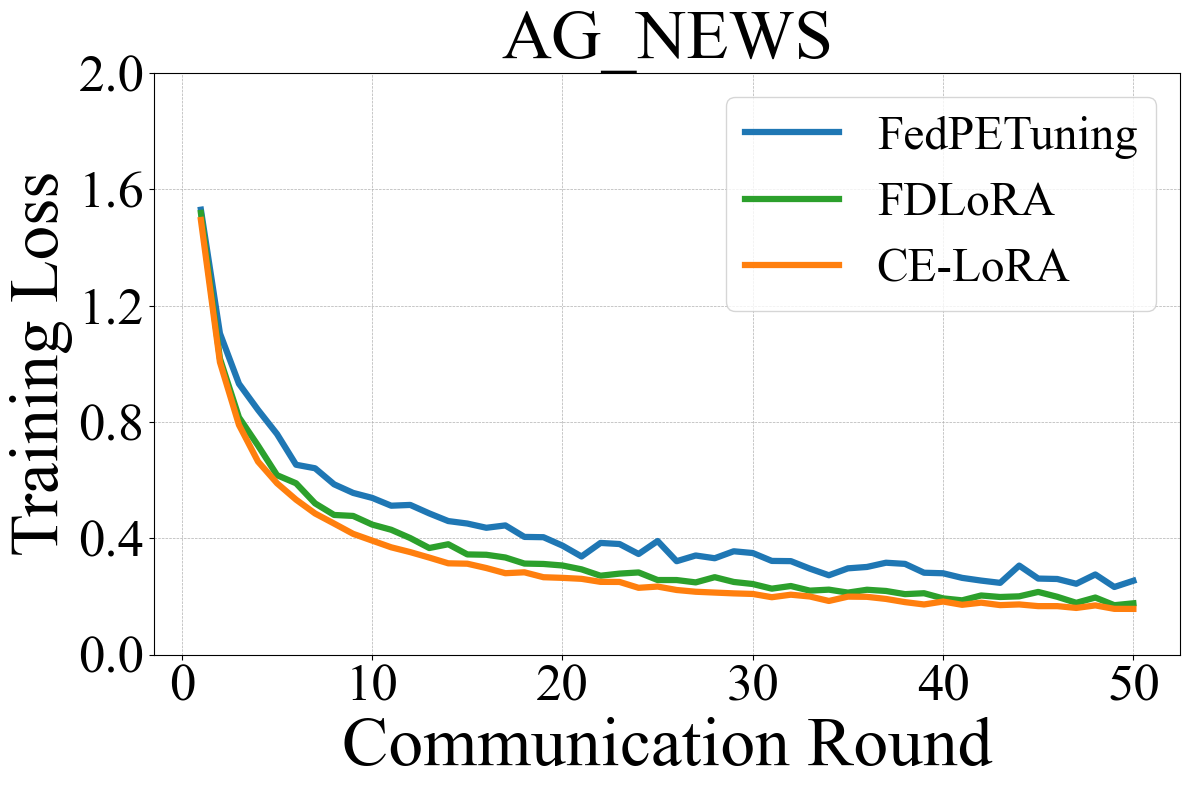}
    \end{subfigure}

    \begin{subfigure}[b]{0.3\textwidth}
        \includegraphics[width=\textwidth]{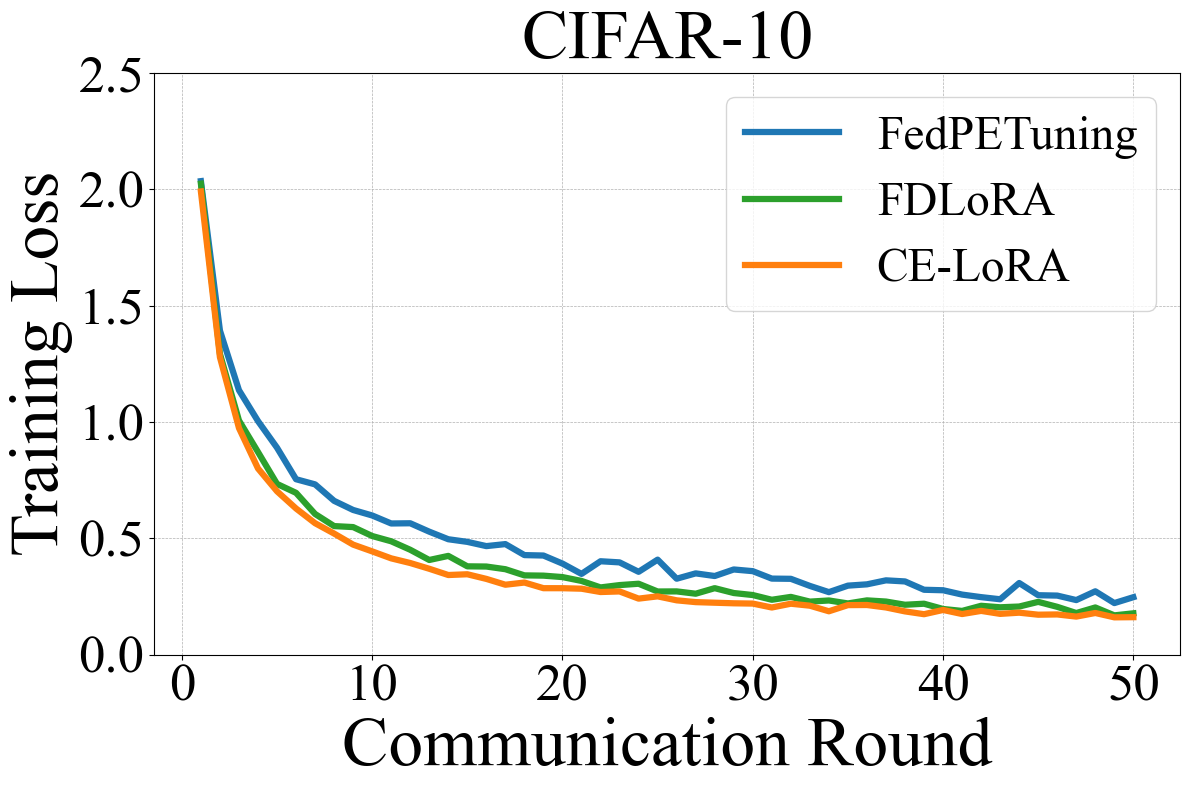}
    \end{subfigure}
    \hfill 
    \begin{subfigure}[b]{0.3\textwidth}
        \includegraphics[width=\textwidth]{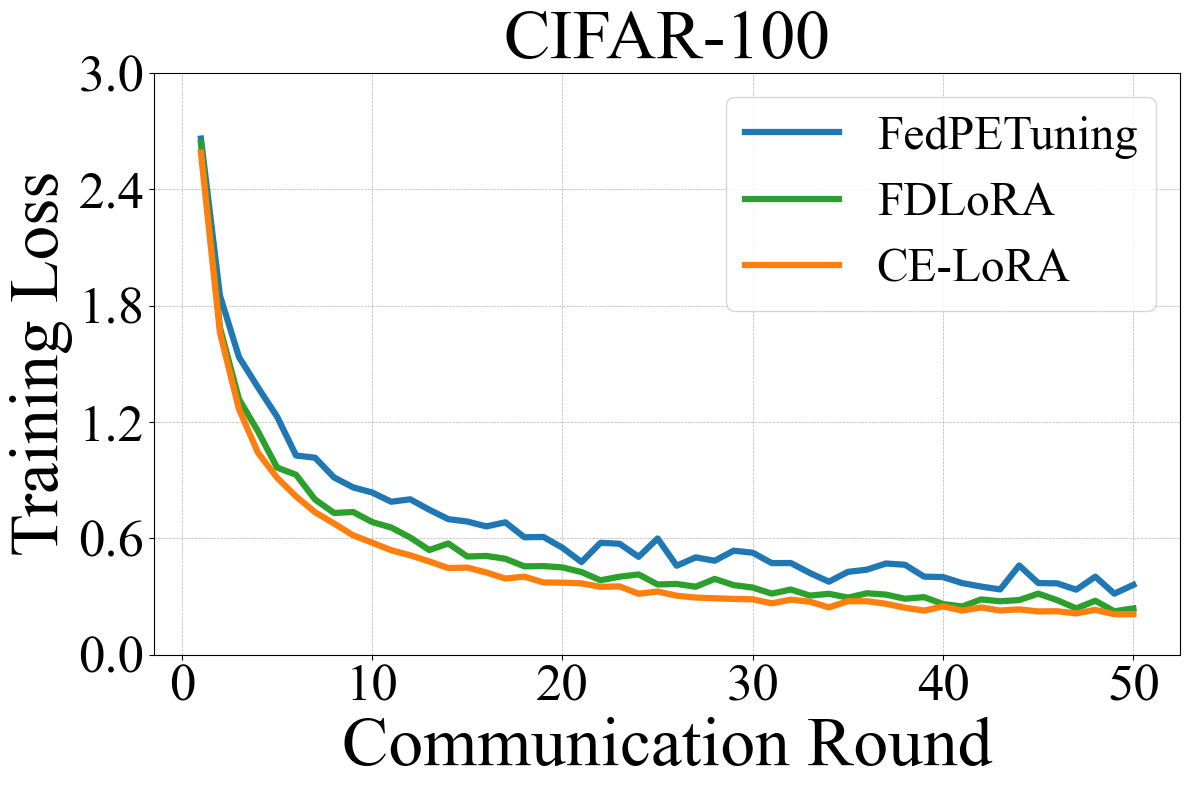}
    \end{subfigure}
    \hfill 
    \begin{subfigure}[b]{0.3\textwidth}
        \includegraphics[width=\textwidth]{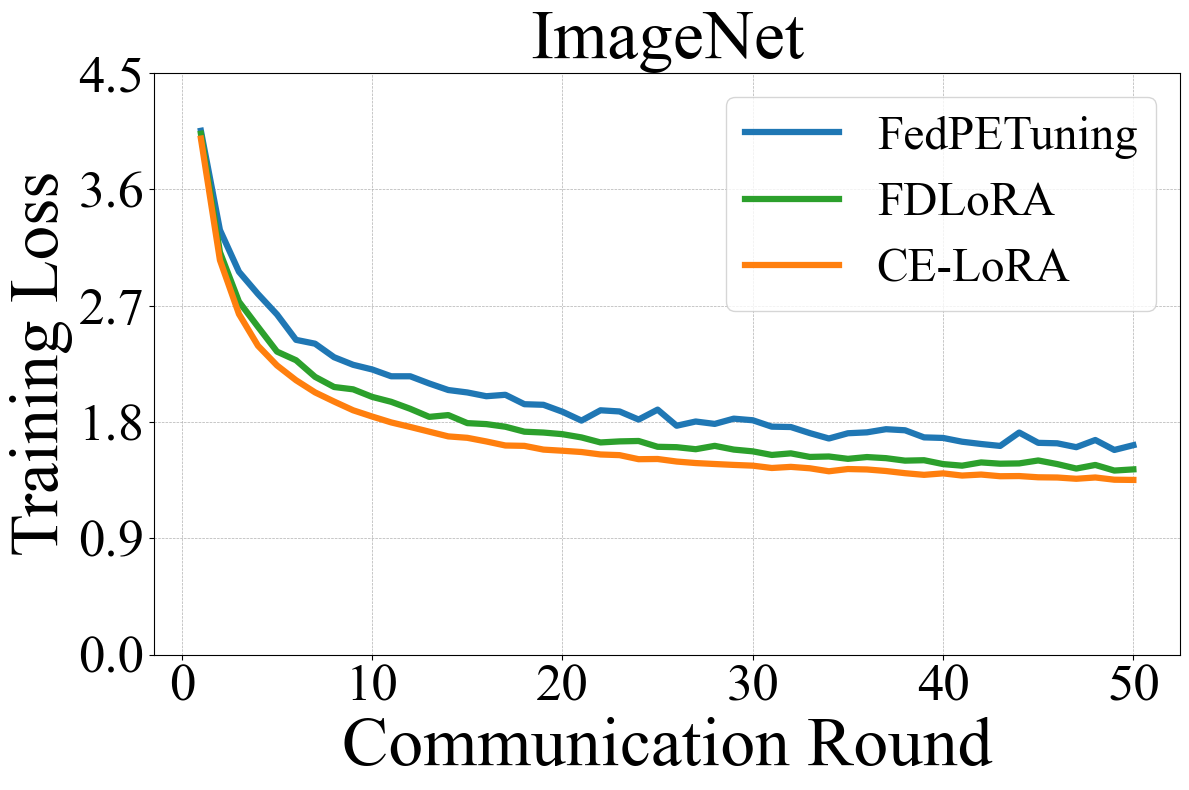}
    \end{subfigure}

\caption{RoBERTa and BLIP-2 fine-tuning convergence comparison of FedPETuning, FDLoRA, and CE-LoRA on the considered benchmark datasets.}
\label{loss}
\vspace{-0.3cm}
\end{figure*}

\subsection{Hyperparameter Analysis}

\begin{figure}[htbp]
    \centering
    \begin{minipage}[b]{0.48\linewidth}
        \centering
        \includegraphics[width=\linewidth]{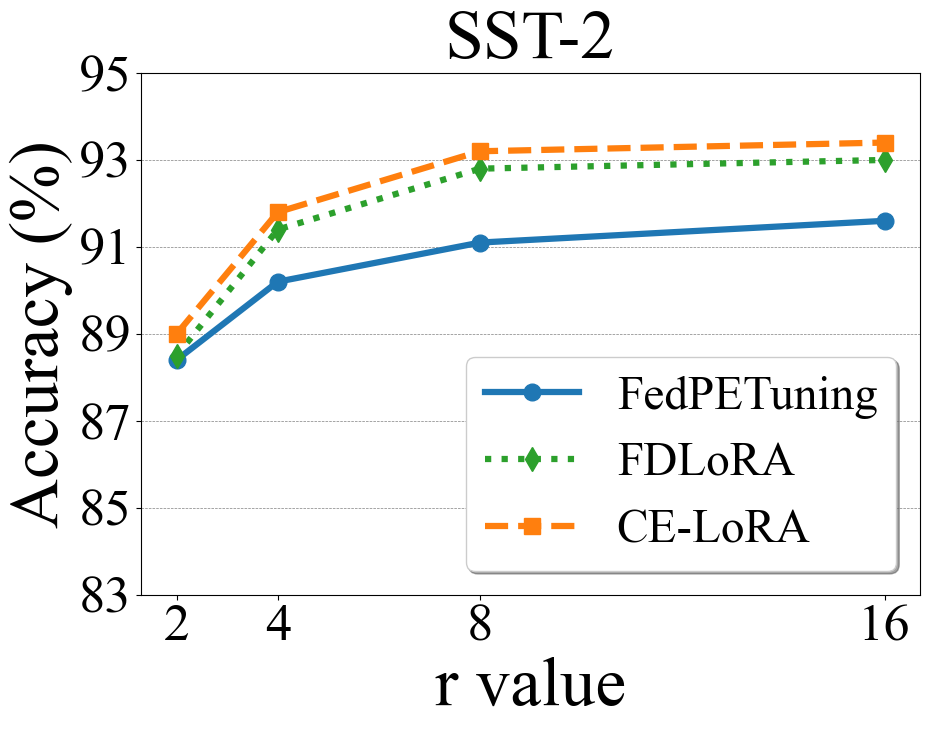}
    \end{minipage}
    \begin{minipage}[b]{0.48\linewidth}
        \centering
        \includegraphics[width=\linewidth]{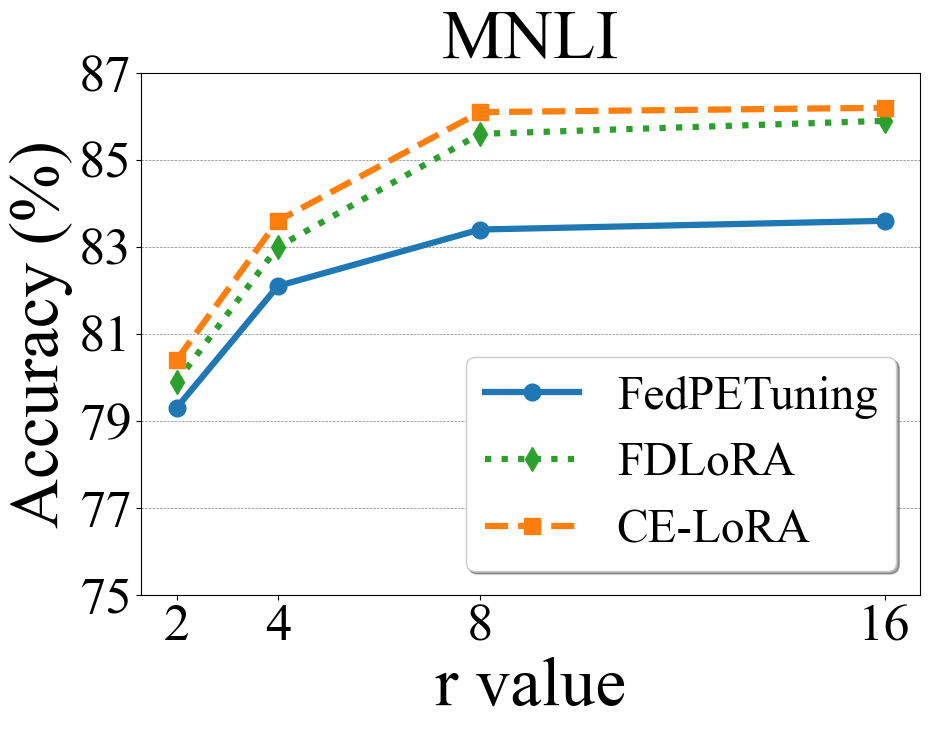}
    \end{minipage}

    \begin{minipage}[b]{0.48\linewidth}
        \centering
        \includegraphics[width=\linewidth]{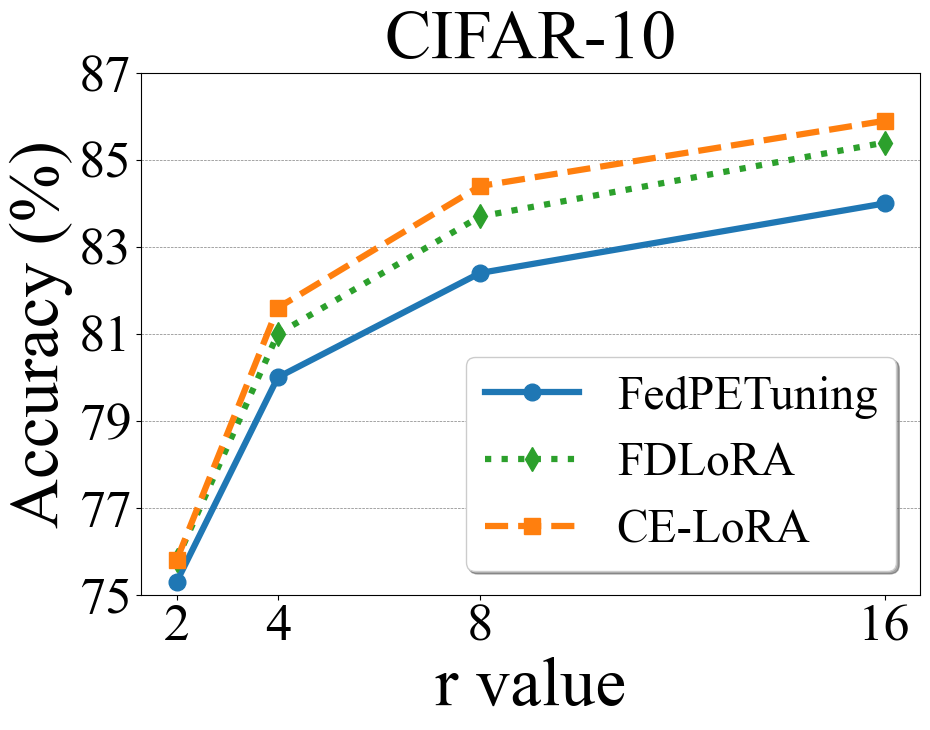}
    \end{minipage}
    \begin{minipage}[b]{0.48\linewidth}
        \centering
        \includegraphics[width=\linewidth]{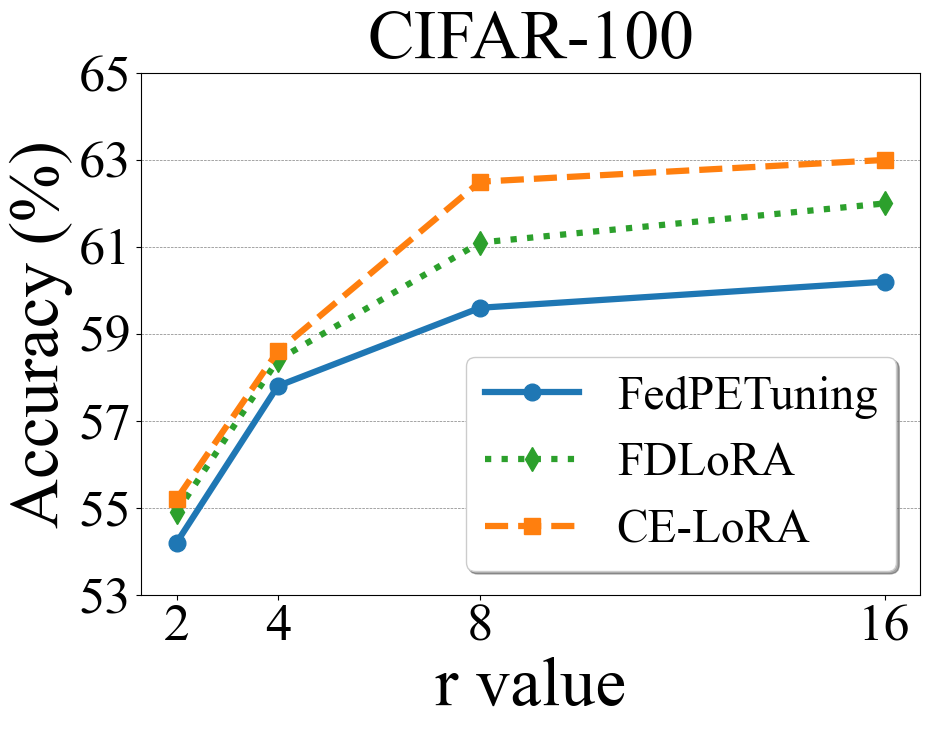}
    \end{minipage}
    
    \caption{Performance comparison of FedPETuning, FDLoRA, and CE-LoRA under different ranks.}
    \label{rank}
\vspace{-0.3cm}    
\end{figure}

In this part, we set up an experiment to observe how performance and communication costs change as the rank $r$ of the low-rank matrices in LoRA changes. We fine-tune RoBERTa on SST-2 and MNLI and BLIP-2 on Cifar10 and Cifar100, respectively. The communication cost increases as $O(r^2)$ with increasing rank $r$. As shown in Figure~\ref{rank}, CE-LoRA achieves a 3.1\% performance gain when increasing $r$ from 2 to 4, while extending $r$ from 8 to 16 yields merely a 0.2\% improvement despite inducing a 16$\times$ larger communication overhead increment compared to the former scaling. There is a trade-off between limited resources and performance, making $r = 4$ or $r = 8$ often a better choice.

\subsection{FL Convergence Analysis}
Figure~\ref{loss} shows the convergence of the FL fine-tuning with respect to the communication round for RoBERTa and BLIP-2 on the benchmark datasets. The results show that CE-LoRA achieves faster convergence and less training loss variation than FedPETuning and FDLoRA. These advantages first stem from CE-LoRA's personalized aggregation strategy, which is based on inter-client data and model similarity, effectively mitigating parameter conflicts of model from different clients. Secondly, the fine-tuning of LoRA component $C$ only in federated approach reduces the parameter number to be learned by FL, which also facilitates faster convergence.


\subsection{Computational Overhead of Personalized Aggregation}

\begin{table}[h]
    \centering
    \fontsize{9}{10}\selectfont
    \setlength{\tabcolsep}{2pt}
    
    \begin{tabular}{c *{4}{>{\centering\arraybackslash}p{1.7cm}}}
        \toprule
        & \multicolumn{2}{c}{RoBERTa} & \multicolumn{2}{c}{BLIP-2} \\
        \cmidrule(lr){2-3} \cmidrule(lr){4-5}
        CPU &   SST-2  &  MNLI  & CIFAR-10 & CIFAR-100 \\
        \midrule
        1  & 14.85  &  15.04 &  20.03 &  20.14  \\
        5  & 4.05  &  4.11   &  5.49  &  5.58   \\
        10 & 2.49  &  2.53  &  3.38  &  3.41  \\
        20 & 1.48  &  1.51 &   2.11 & 2.12  \\
        \bottomrule
    \end{tabular}

    \caption{Computational Overhead of CE-LoRA Aggregation with 100 Clients (seconds).}
    \label{tab:ce_lora_overhead}
\vspace{-0.3cm}
\end{table}

The computational complexity of the model similarity evaluation presented in is $O(m^2)$ where $m$ is the number of clients. However, since the transmitted LoRA component $C$ is compact, the similarity computation is lightweight. The overall computation can be further speed up by parallel computing. As shown in Table~\ref{tab:ce_lora_overhead}, we evaluate the time cost of pairwise model similarity computation across 100 clients under varying levels of CPU parallelism with 1, 5, 10 and 20 CPUs, respectively. For even larger number of clients, using hierarchical structure such as the client-edge-cloud aggregation hierarchy~\cite{liu2022hierarchical} and client sampling~\cite{chen2020optimal} can handle the scalability problem.

\section{Conclusion}

This paper introduces the CE-LoRA method, which leverages triple LoRA factorization and personalized global model aggregation to address the challenges of high communication overhead and suboptimal performance on heterogeneous data in federated learning environments. Experimental results demonstrate that CE-LoRA not only reduces communication costs but also significantly enhances model performance on non-IID data, particularly in environments with highly heterogeneous data distributions. Moreover, it excels in protecting data privacy, effectively resisting gradient-based data reconstruction attacks. Our work primarily addresses data heterogeneity, with future research extending CE-LoRA to task and model heterogeneity to broaden its applicability in federated learning.


\bibliographystyle{IEEEtran}
\bibliography{main.bib}

\begin{IEEEbiography}[{\includegraphics[width=1in,height=1.25in,clip,keepaspectratio]{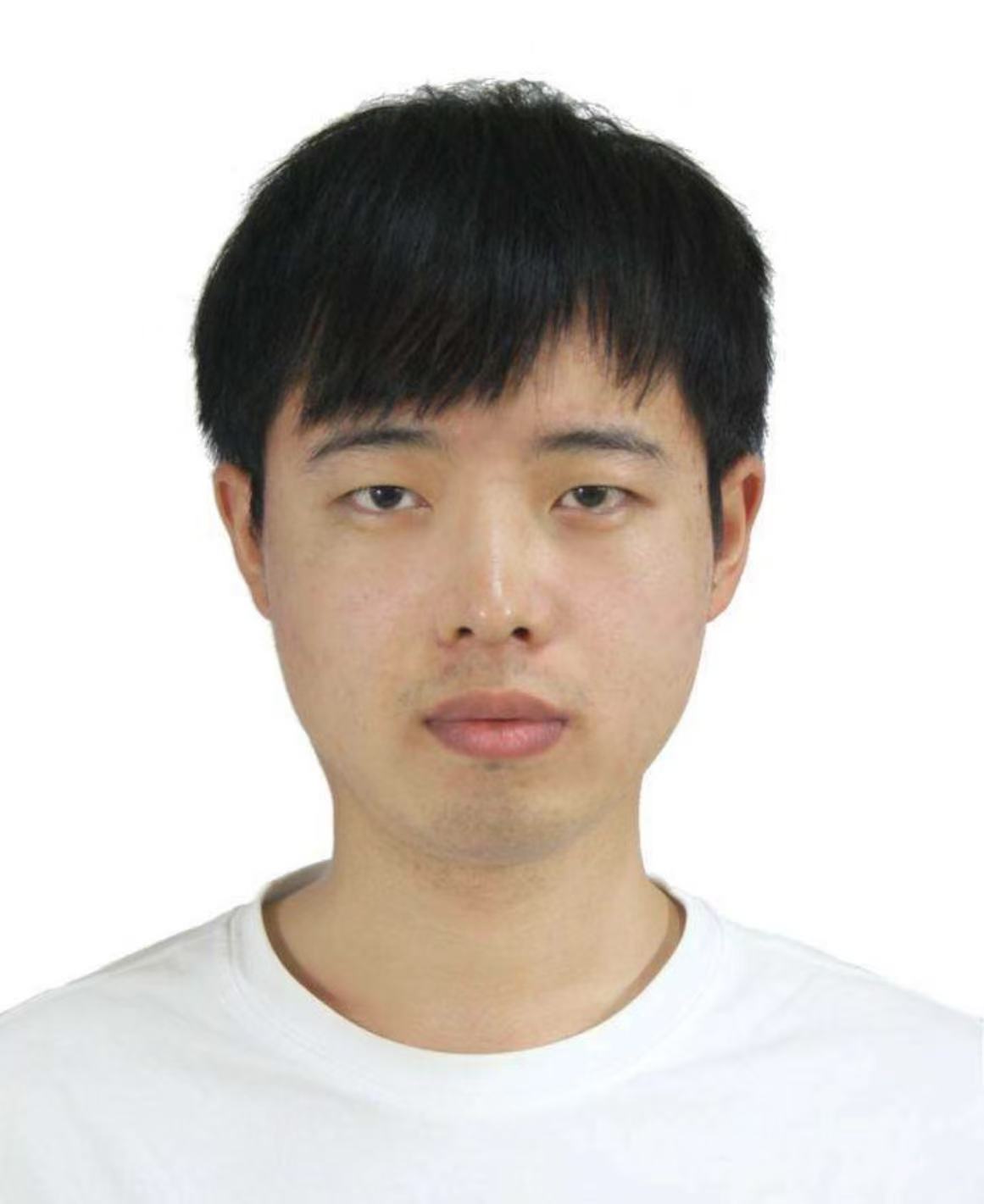}}]{Yongle Li}
is currently a PhD candidate at the School of Computer and Information, Hefei University of Technology, China, under the supervision of Prof. Richang Hong (since 2023). His research interests include fine-tuning pre-trained foundation models and federated learning.
\end{IEEEbiography}
\vspace{-1cm}
\begin{IEEEbiography}[{\includegraphics[width=1in,height=1.25in,clip,keepaspectratio]{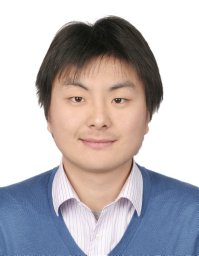}}]{Bo Liu}
received the Ph.D. degree in computer science from Rutgers, The State University of New Jersey, in 2018. He currently is a faculty member at Hefei University of Technology, Hefei, China. Prior to this role, He held the positions of Staff Data Scientist and Senior Research Scientist at Walmart Global Tech and JD.com Silicon Valley Research Center, respectively. His areas of expertise and research interests encompass machine learning, computer vision, and data science.
\end{IEEEbiography}
\vspace{-1cm}
\begin{IEEEbiography}[{\includegraphics[width=1in,height=1.25in,clip,keepaspectratio]{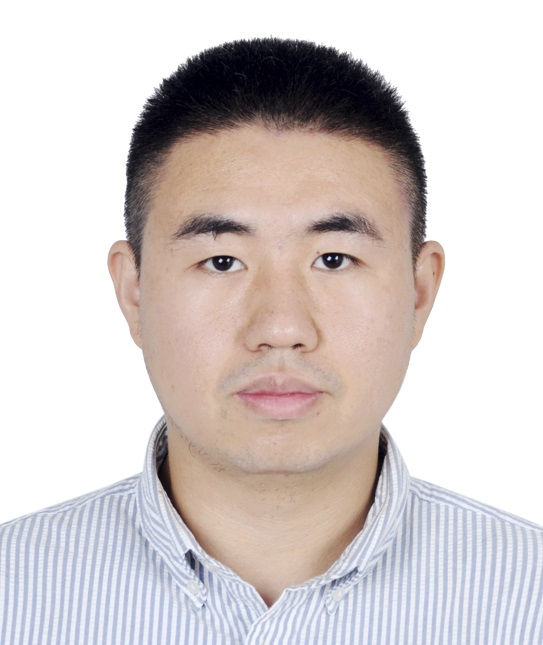}}]{Sheng Huang}
(Member, IEEE) received the B.Eng. and Ph.D. degrees from Chongqing University, Chongqing, China, in 2010 and 2015, respectively. He was a visiting Ph.D. Student with the Department of Computer Science, Rutgers University, New Brunswick, NJ, USA, from 2012 to 2014. He is currently a full Professor with the School of Big Data and Software Engineering, Chongqing University, China. His research interests include pattern recognition, computer vision and medical image analysis.
\end{IEEEbiography}
\vspace{-1cm}
\begin{IEEEbiography}[{\includegraphics[width=1in,height=1.25in,clip,keepaspectratio]{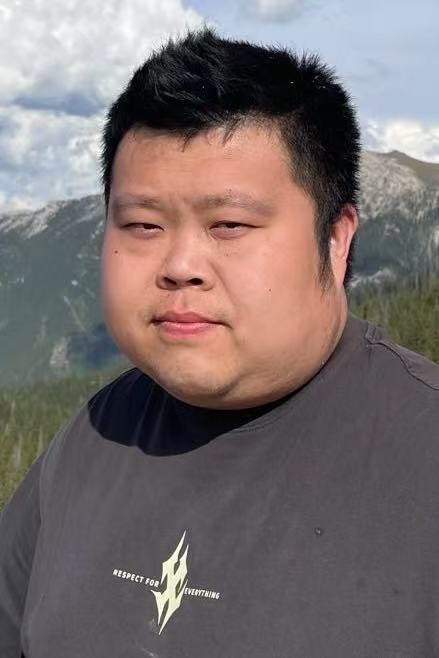}}]{Zheng Zhang}
Graduated from Tsinghua University, he has over 7 years of professional experience, consistently working in start-up companies. They possess extensive research and industrial expertise in deep learning, natural language processing (NLP) technologies, and large-scale models. he has led the development of initiatives such as the Atom Large Language Model, demonstrating profound contributions to both academic research and real-world applications in these fields.
\end{IEEEbiography}
\vspace{-1cm}
\begin{IEEEbiography}[{\includegraphics[width=1in,height=1.25in,clip,keepaspectratio]{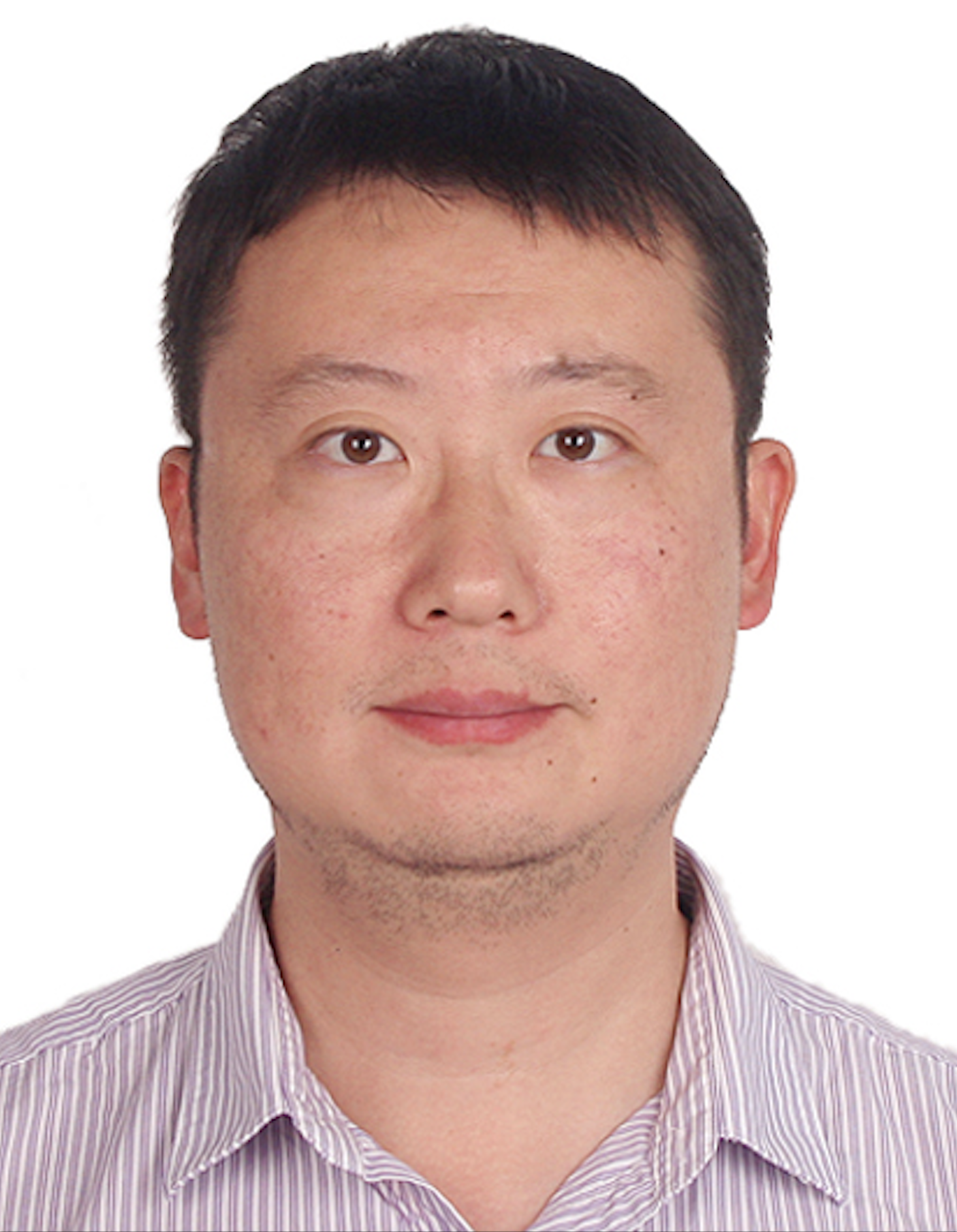}}]{XiaoTong Yuan}
(Member, IEEE) received the B.A. in computer science from Nanjing University of Posts and Telecommunications in 2002, the M.E. in electrical engineering from Shanghai Jiao-Tong University in 2005, and the Ph.D. in pattern recognition from the Chinese Academy of Sciences in 2009. He then worked as a Postdoctoral Research Associate at the National University of Singapore, Rutgers University, and Cornell University. In 2013, he joined Nanjing University of Information Science and Technology, where he is currently a Professor of computer science. His main research interests include machine learning, optimization, and computer vision.
\end{IEEEbiography}
\vspace{-1cm}
\begin{IEEEbiography}[{\includegraphics[width=1in,height=1.25in,clip,keepaspectratio]{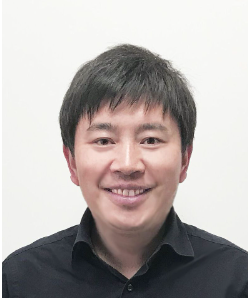}}]{Richang Hong}
is a professor and PhD supervisor at Hefei University of Technology. He currently
serves as the Dean of the School of Computer and Information (School of Artificial Intelligence) and the School of Software at Hefei University of Technology. He is also the Deputy Director of the Data Space Research Institute at the Hefei Comprehensive National Science Center and the President of the Anhui Artificial Intelligence Society. His research focuses on artificial intelligence-related fields, with over 300 high-level papers published and more than 20,000 citations. He serves as a Steering Committee member of the International Conference on Multimedia Modeling and as an editorial board member of six international journals, including IEEE TBD, IEEE TMM, and ACM TOMM. He has led various national projects, including the Ministry of Science and Technology’s 863 Program, the Ministry’s Key Research and
Development Program, and projects funded by the National Natural Science Foundation of China, such as the Excellent Young Scientists Fund and key foundation projects. His research achievements have been recognized with the National Natural Science Award (Second Prize, 2015), Anhui Provincial Natural Science Award (First Prize, 2017), and Anhui Provincial Science and Technology Progress Award (First Prize, 2020). In teaching, he received the Anhui Provincial Teaching Achievement Award (First Prize, 2022) and the National Teaching Achievement Award (First Prize for Postgraduate Education, 2023). He was awarded the Anhui Province Youth May Fourth
Medal in 2019 and was selected for the national leading talent program.
\end{IEEEbiography}

\vspace{11pt}

\vfill

\end{document}